\documentclass[nohyperref]{article}

\usepackage{microtype}
\usepackage{graphicx}
\usepackage{subfigure}
\usepackage{booktabs} % for professional tables
\usepackage{comment}

\usepackage{hyperref}

\usepackage[accepted]{icml2023}

\usepackage{amsmath}
\usepackage{amssymb}
\usepackage{mathtools}
\usepackage{amsthm}

\usepackage{lipsum} 

\usepackage[capitalize,noabbrev]{cleveref}
\newcommand{\R}{\mathbb{R}}
\newcommand{\Rd}{\R^d}
\newcommand{\supp}{\mathrm{Spt}}
\theoremstyle{plain}
\newtheorem{theorem}{Theorem}[section]
\newtheorem{proposition}[theorem]{Proposition}
\newtheorem{lemma}[theorem]{Lemma}
\newtheorem{corollary}[theorem]{Corollary}
\theoremstyle{definition}
\newtheorem{definition}[theorem]{Definition}

\theoremstyle{remark}

\usepackage[textsize=tiny]{todonotes}

\newcommand{\Id}{\mathrm{I}_d}
\newcommand*\diff{\mathop{}\!\mathrm{d}}
\newcommand*\simiid{\sim_{\mathrm{i.i.d}}}
\newcommand*{\eqdef}{\vcentcolon =}
\newcommand*{\jac}{\mathrm{Jac}}

\def\*#1{\mathbf{#1}}
\def\emp*#1{\hat{#1}_n}
\def\emp*#1{\hat{#1}_n}

\usepackage[inline]{enumitem}
\icmltitlerunning{The Monge Gap}
\begin{document}

\twocolumn[
\icmltitle{The Monge Gap: A Regularizer to Learn All Transport Maps}




\begin{icmlauthorlist}
\icmlauthor{Théo Uscidda}{ensae}
\icmlauthor{Marco Cuturi}{ensae,apple}
\end{icmlauthorlist}

\icmlaffiliation{ensae}{CREST, ENSAE}
\icmlaffiliation{apple}{Apple}

\icmlcorrespondingauthor{Théo Uscidda}{theo.uscidda@ensae.fr}
\icmlcorrespondingauthor{Marco Cuturi}{cuturi@apple.com}

\icmlkeywords{Machine Learning, ICML}

\vskip 0.3in
]



\printAffiliationsAndNotice{}  % leave blank if no need to mention equal contribution

\begin{abstract}
Optimal transport (OT) theory has been been used in machine learning to study and characterize maps that can push-forward efficiently a probability measure onto another.
Recent works have drawn inspiration from~\citeauthor{Brenier1987}'s theorem, which states that when the ground cost is the squared-Euclidean distance, the ``best'' map to morph a continuous measure in $\mathcal{P}(\Rd)$ into another must be the gradient of a convex function.
To exploit that result, \citet{makkuva2020optimal, korotin2020wasserstein} consider maps $T=\nabla f_\theta$, where $f_\theta$ is an input convex neural network (ICNN), as defined by \citet{amos2017input}, and fit $\theta$ with SGD using samples.
Despite their mathematical elegance, fitting OT maps with ICNNs raises many challenges, due notably to the many constraints imposed on $\theta$; the need to approximate the conjugate of $f_\theta$; or the limitation that they only work for the squared-Euclidean cost. More generally, we question the relevance of using \citeauthor{Brenier1987}'s result, which only applies to densities, to constrain the architecture of candidate maps fitted on samples.
Motivated by these limitations, we propose a radically different approach to estimating OT maps:
Given a cost $c$ and a reference measure $\rho$, we introduce a regularizer, the Monge gap $\mathcal{M}^c_{\rho}(T)$ of a map $T$. That gap quantifies how far a map $T$ deviates from the ideal properties we expect from a $c$-OT map. In practice, we drop all architecture requirements for $T$ and simply minimize a distance (e.g., the Sinkhorn divergence) between $T\sharp\mu$ and $\nu$, regularized by $\mathcal{M}^c_\rho(T)$. We study $\mathcal{M}^c_{\rho}$, and show how our simple pipeline outperforms significantly other baselines in practice.
\end{abstract}

\section{Introduction}\label{sec:intro}
At the core of many machine learning challenges lies the problem of learning a map $T:\mathbb{R}^d\rightarrow \mathbb{R}^d$ that is able to push-forward a probability measure $\mu$ into another, $\nu$, i.e., $T\sharp \mu=\nu$.
If one were given paired samples $(\*x_i,\*y_i)$, the task would amount to a simple regression, easily solved by minimizing an averaged risk $c(T(\*x_i),\*y_i)$. In many applications, however, only unmatched samples $(\*x_1,\dots,\*x_n)$ from $\mu$ and $(\*y_1,\dots,\*y_m)$ from $\nu$ are provided, requiring a distributional approach to estimate $T$. When the input measure $\mu$ is simple and closed-form (e.g. Gaussian, or uniform), likelihood-based methods can be used, notably normalizing flows~\cite{rezende2015variational}, GANs~\citep{goodfellow2014generative} or even diffusion models~\cite{song2020score}.

\textbf{Optimal Transport and the Brenier Story. } When both measures are complex and can only be accessed through samples, finding a good map $T$ poses extra challenges. This is the case, e.g., in domain adaptation~\citep{Courty-2016-Optimal,Courty-2017-Joint} or in genomics~\citep{schiebinger2019}. Optimal transport (OT) theory~\citep{santambrogio2015optimal} has emerged as a prime contender for that task~\citep{Peyre2019computational}. We focus in this work on neural OT solvers, where $T$ is parameterized as a neural network. That area has been largely shaped by~\citeauthor{Brenier1987}'s theorem, which states that when the cost is the squared-Euclidean distance, OT maps should follow the gradients of a convex potential. Leveraging that result,\citet{makkuva2020optimal,korotin2020wasserstein} provided a blueprint to use input convex neural networks (ICNN) for OT estimation, which was later exploited in various applications, notably genomics~\citet{bunne2021learning}.

\textbf{On the limitations of ICNNs for OT. } While the theory motivating ICNN solvers for OT is compelling, their practical implementation runs into many challenges~\citep{korotin2021do}: some of their parameters must be non-negative, initialization them, although the subject of ongoing research~\citep{korotin2020wasserstein,bunne2022supervised}, is still poorly understood, and training them requires approximating a convex conjugate with a min-max formulation~\cite{amos2022amortizing}. On a more fundamental level, the ICNN approach may not be as sound as it seems: while \citet{Brenier1987}'s argument is valid when the input measure $\mu$ is a density, that result does not hold for sample measures. One might therefore question the relevance of imposing the double requirement that a candidate map be the \textit{gradient} of a \textit{convex} potential. For general costs $c$, these requirements are equivalent to a $c$-concavity constraint that is even more intractable when trying to generalize ICNNs to other costs~\citep{rezende2021implicit,cohen2021riemannian}. We question the need for such constraints, as also done, for instance for score functions in score-based models~\citep{saremi2019approximating}.

\textbf{Contributions. } We propose a new approach to estimate OT maps, sturdy and generic enough to work for any cost $c$.
\begin{itemize}[leftmargin=.3cm,itemsep=0cm,topsep=0cm,parsep=1pt]
\item Rather than imposing architecture choices to mimic OT maps, we make no assumption on $T$ and, instead, introduce a \textit{regularizer} which quantifies whether $T$ agrees with the theoretical properties needed for $T$ to be an OT map.
\item The \textit{Monge gap} regularizer $\mathcal{M}^c_\rho$ uses a \textit{reference} measure $\rho$ (that need not be necesseraly equal to $\mu$), and is the difference between the expectation of $c(X,T(X))$, $X\sim\rho$, and the $c$-Wasserstein distance between $\rho$ and $T\sharp \rho$.
\item We show that the Monge gap characterizes the optimality of a map $T$ between $\mu$ and $\nu$. More formally, when $T\sharp\mu=\nu$ and the support $\supp(\mu)\subset \supp(\rho)$, we show that $\mathcal{M}^c_\rho(T)=0$ iff $T$ is an optimal map. 
\item We show that $\mathcal{M}^c_\rho$ is convex when $c(\cdot,\cdot)=\|\cdot-\cdot\|^2_2$, a property which is \textit{still} valid when using a Sinkhorn finite-sample estimator for the $2$-Wasserstein distance.
\item We propose two learning procedures to estimate Monge maps using the Monge gap: (i) for general costs $c$, we simply add the Monge Gap of a vector field $T$ to a fitting loss measuring the difference between $T\sharp\mu$ and the true target distribution $\nu$ and (ii) when the cost satisfies the twist condition, we take advantage of the structure induced by such costs on the optimal map, and propose instead to directly parameterize the gradient of the potential. 
\item We provide ample evidence on toy data, synthetic benchmarks~\citep{korotin2021do} and single-cell data that our regularized approach outperforms both ICNNs and vanilla MLPs, but also works for other more exotic costs.
\end{itemize}
\section{Background on optimal transport}

\textbf{Monge and Kantorovich formulation.} We consider throughout this work a compact subset $\Omega \subset \mathbb{R}^d$, a continuous cost function $c : \Omega \times \Omega \rightarrow \mathbb{R}$ and two probability distributions $\mu$, $\nu \in \mathcal{P}(\Omega)$. The notation $\mu \in \mathcal{P}(\Omega)$, $\mu \ll \mathcal{L}_d$ means that $\mu$ is absolutely continuous w.r.t. the Lebesgue measure. %For a measurable map $T$, $T \sharp \mu $ is the pushforward of $\mu$ by the map $T$; for all measurable sets $A \subset \Omega$, $T \sharp \mu (A) = \mu(T^{-1}(A))$. 
The \citeauthor{Monge1781} problem consists of finding, among all map $T: \Omega \rightarrow \Omega$ that push-forward $\mu$ onto $\nu$, that which minimizes the averaged displacement cost:
\begin{equation}
\label{eq:monge-problem}
W_c(\mu, \nu) \eqdef \inf_{T\sharp\mu=\nu} \int_\Omega c(\*x, T(\*x)) \diff\mu(\*x)\,.
\end{equation}
We call any solution to \eqref{eq:monge-problem} a $c$-OT map between $\mu$ and $\nu$. Solving this problem is difficult: the constraint set is not convex and can even be empty, when, for instance, $\mu$ is discrete and $\nu \ll \mathcal{L}_d$. Instead of transport maps, the \citet{kantorovich1942transfer} formulation of OT seeks for couplings $\pi \in \Pi(\mu, \nu)$, i.e., probability measures supported on $\Omega \times \Omega$ that have $\mu$ and $\nu$ as respective marginals:
\begin{equation}
\label{eq:kantorovich-problem}
W_c(\mu, \nu) \eqdef \min_{\pi \in \Pi(\mu, \nu)} \iint_{\Omega\times \Omega} c(\*x, \*y) \diff\pi(\*x, \*y)\,.
\end{equation}

An optimal coupling $\pi^\star$ always exists. When Problem \eqref{eq:monge-problem} is feasible, both formulations coincide in the sense that the optimal coupling will be concentrated on the graph of $T^\star$, namely\  $\pi^\star = (\mathrm{Id}, T^\star)\sharp\mu$.

\textbf{Primal-dual relationship. } For any $\varphi : \Omega \rightarrow \mathbb{R}$, writing $\varphi^c : \*y \in\Omega \mapsto \inf_\*x c(\*x, \*y) - \varphi(\*x)$ its $c$-transform, one can derive the Kantorovich dual:
\begin{equation}
\label{kantorovich-dual}
W_c(\mu, \nu) = \min_{\varphi \,:\, \Omega \rightarrow \mathbb{R}} \int_\Omega \varphi \diff \mu + \int_\Omega \varphi^c \diff \nu \,.
\end{equation}

Taking an optimal potential $\varphi^\star$ (also called a Kantorovich potential, which always exists under our assumptions on $\Omega$ and $c$) and an optimal coupling $\pi^\star$, the complementary slackness reads: $\forall (\*x_0, \*y_0) \in \mathrm{Spt}(\pi^\star), \varphi^\star(\*x_0) + \varphi^{\star, c}(\*y_0) = c(\*x_0, \*y_0)$. 
Assume that $\varphi^\star$ is differentiable at $\*x_0$, which is true under mild assumptions, and that $c$ is differentiable w.r.t.\ the first variable. Exploiting the definition of $\varphi^{\star, c}$:
\begin{equation}
\label{eq:support-pi-star}
(\*x_0, \*y_0) \in \mathrm{Spt}(\pi^*) \Leftrightarrow \nabla \varphi^\star(\*x_0) = \nabla_1 c(\*x_0, \*y_0)\,.
\end{equation}
From there, if $c$ satisfies the so-called twist condition \citep[Definition 1.16]{santambrogio2015optimal}, namely for all $\*x$, $\nabla_1 c(\*x, \cdot)$ is injective, the optimal map reads: 
\begin{equation}
\label{eq:optimal-map}
T^\star : \*x \mapsto \nabla_1 c(\*x, \cdot)^{-1} \circ \nabla \varphi^\star(\*x)\,.
\end{equation}
Indeed, thanks to Equation \ref{eq:support-pi-star} and the inversibility assumption, $\pi^
\star$ is concentrated on the graph of this map. When $c(\*x, \*y) = h(\*x - \*y)$ with $h : \Omega \rightarrow \mathbb{R}$ is strictly convex, the differentiability assumption on $c$ can be relaxed. While $h$ is only subdifferentiable (as a convex function \citep[Section 23]{rockafellar1970convex}), its subgradient, which is a multi-valued map, can be inverted and is uni-valued. Indeed, one has $(\partial h)^{-1}(\*x) = \{\nabla h^*(\*x)\}$, with $h^*$ the convex conjugate of $h$ \citep[Box 1.12]{santambrogio2015optimal}. In that specific case:
\begin{equation}
\label{eq:optimal-map-convex-costs}
T^\star : \*x \mapsto \*x - \nabla h^* \circ \nabla \varphi^\star(\*x)\,.
\end{equation}
In particular, when $h = \frac{1}{2} \| \cdot  \|_2^2$  one recovers the \citet{Brenier1987} Theorem: $T^\star = \mathrm{Id} - \varphi^\star = \nabla f^\star$ where $f^\star \eqdef \tfrac{1}{2} \|\cdot\|_2^2 - \varphi^\star$ can be shown to be convex.

\paragraph{Entropic regularization.}

When both $\mu$ and $\nu$ are instantiated as samples, as usual in a
machine learning context, the~\citet{kantorovich1942transfer} Problem~\eqref{eq:kantorovich-problem} translates to a linear program, whose objective can be smoothed out using entropic regularization \citep{cuturi2013sinkhorn}. For empirical measures $ \emp*\mu = \frac{1}{n} \sum_{i=1}^n \delta_{\*x_i}$, $\emp*\nu = \frac{1}{n} \sum_{j=1}^n \delta_{\*y_j}$ and $\varepsilon > 0$, we form $\*C = \left[ c(\*x_i, \*y_j) \right]_{ij}$ and set:
\begin{equation}
\label{eq:entropicOT}
W_{c, \varepsilon}(\emp*\mu, \emp*\nu) \eqdef \min_{\*P \in U_n} \langle \*P, \*C \rangle - \varepsilon H(\*P)\,,
\end{equation}
where $U_n = \{\*P \in \mathbb{R}^{n \times m}_+, \*P \*1_n = \tfrac{1}{n} \*1_n, \*P^T \*1_n = \tfrac{1}{n} \*1_n \}$
is the Birkhoff polytope and $H(\*P) = - \sum_{i,j=1}^n \*P_{ij} \log(\*P_{ij})$ the entropy. As $\varepsilon$ goes to $0$, one recovers the classical OT problem, namely $W_{c, 0} = W_c$. In addition to resulting in better computational and statistical performance~\citep{genevay2019sample,mena2019statistical,chizat2020faster}, entropic regularization also results in a strongly convex problem, with a unique solution, making $W_{c, \varepsilon}$ differentiable everywhere in its inputs via \citep{danskin2012theory}'s theorem. Besides, one can define
% entropic regularization it introduces a bias since even when $c$ is a distance, $\argmin_\mu W_{c, \varepsilon}(\mu, \nu) \ne \nu$ in general. 
the Sinkhorn divergence $S_{c, \varepsilon}(\mu, \nu) \eqdef W_{c, \varepsilon}(\mu, \nu) - \frac{1}{2} \left( W_{c, \varepsilon}(\mu, \mu) + W_{c, \varepsilon}(\nu, \nu) \right)$ \citep{ramdas2017wasserstein, feydy2018interpolating, salimans2018improving, genevay2018} which is, under some assumptions on $c$ (see \citet[Theorem 1]{feydy2018interpolating}), a valid non-negative discrepancy measure between probability distributions. The quadratic cost satisfies theses assumptions and we we note $W_{\ell_2^2 , \varepsilon}$ and $S_{\ell_2^2 , \varepsilon}$ in that case.

\section{The Monge Gap}
We introduce in this section the Monge gap, a regularizer to estimate optimal transport maps with any ground cost $c$.  
 \begin{definition}[The Monge Gap]
\label{def:convexity-regularization}
Given a cost $c$ and a reference measure $\rho \in \mathcal{P}$, the Monge gap of a vector field $T : \Omega \rightarrow \Omega$ is defined as:
\begin{align}
\begin{split}
\mathcal{M}^c_{\rho}(T) \eqdef & \int_\Omega c(\*x, T(\*x)) \diff \rho(\*x) - W_c(\rho, T \sharp \rho)\,.
\end{split}
\end{align}
\end{definition}

By definition of Eq.~\eqref{eq:monge-problem}, the Monge problem between $\rho$ and $T \sharp \rho$ is feasible for any measure, notably discrete, since there exists at least one map, $T$ itself, that satisfies the push-forward constraint. With this in mind, because the Monge gap is simply the optimality gap of the Monge problem, one can deduce immediately the following properties: 
\begin{itemize}[leftmargin=.3cm,itemsep=.0cm,topsep=0cm]
\item For any vector field $T$, $\mathcal{M}^c_{\rho}(T) \geq 0$.
\item $T$ is a $c$-OT map between $\rho$ and $T\sharp\rho \Leftrightarrow \mathcal{M}^c_{\rho}(T)=0$.
\end{itemize}

\begin{figure}[h]
     \centering
     \includegraphics[width=.75\columnwidth]{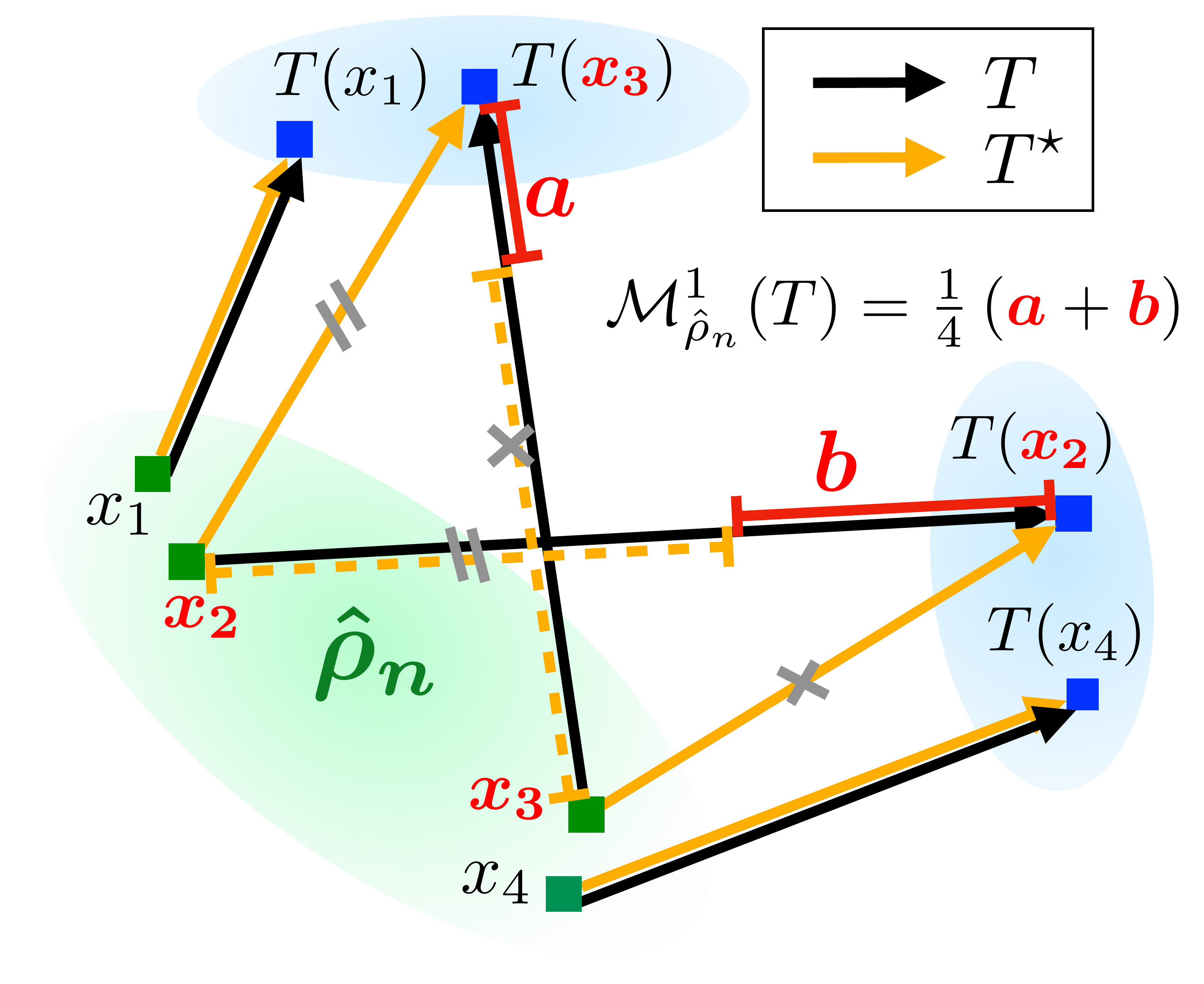}
     \caption{
     Sketch of the Monge Gap $\mathcal{M}_{\hat\rho_n}^1(T)$ instantiated with the euclidean cost $c(\cdot, \cdot) = \|\cdot - \cdot\|_2$, where $\hat\rho_n$ is a discrete measure supported on four points. Because the OT map $T^\star$ between $\hat\rho_n$ and $T\sharp{\hat\rho_n}$ does not coincide with $T$ (notably on on points ${\color{red}x_2,x_3}$), the Monge gap here is positive, and equal to differences in lengths that amount to $({\color{red}a}+{\color{red}b})/4$ in the plot.
     }
     \label{fig:sketch}
\end{figure}

Intuitively, the Monge gap $\mathcal{M}^c_{\rho}$ measures the gap between the cost incurred when moving from $\rho$ to $T\sharp\rho$ using $T$, to the optimal one (not necessarily $T$) realized by a $c$-OT map $T^\star$. See Figure~\ref{fig:sketch} for a simple illustration.

\subsection{Estimation from Samples.}
In practice, we estimate the Monge gap using i.i.d. samples $\*x_1, ..., \*x_n$ from $\rho$. Given empirical measures $\emp*\rho \eqdef \frac{1}{n} \sum_{i=1}^n \delta_{\*x_i}$ and $T\sharp\emp*\rho = \frac{1}{n} \sum_{i=1}^n \delta_{T(\*x_i)}$, we can simply consider the plug-in estimator $\mathcal{M}^c_{\emp*\rho}(T)$. Under mild assumptions guarantying that $T \sharp \emp*\rho \rightarrow T \sharp \rho$ in law, we show that $\mathcal{M}_{\emp*\rho}^c$ is a consistent estimator of $\mathcal{M}_{\rho}^c$.
\begin{lemma}[Consistency]\label{lem:consistency}
\label{lem:consistency}
Provided that $T$ is continuous, it almost surely holds: 
\begin{align}
% \lim_{n \to +\infty} \mathcal{M}^c_{\emp*\rho, \varepsilon_n}(F) = 
\lim_{n \to +\infty} \mathcal{M}^c_{\emp*\rho}(T) = \mathcal{M}^c_{\rho}(T)
\end{align}
\end{lemma}
\begin{proof}
For the RHS, let $\*x_1, ..., \*x_n \simiid \rho$, then almost surely, $T \sharp \emp* \rho \rightarrow T \sharp \rho$ in law. Indeed, if $g : \Omega \rightarrow \mathbb{R}$ is bounded and continuous, as well as $g \circ T$, then:
\begin{align*}
\int g \diff T \sharp \emp*\rho = \int g \circ T \diff \emp*\rho \rightarrow \int g \circ T \diff \rho = \int g \diff T \sharp \rho
\end{align*}
since, almost surely, $\emp*\rho \rightarrow \rho$ in law. Then, since $c$ is continuous and $\Omega$ is compact, one has $W_c(\emp*\rho, T\sharp\emp*\rho) \rightarrow W_c(\rho, T\sharp\rho)$ \citep[Theorem 1.51]{santambrogio2015optimal}, hence almost surely $\mathcal{M}^c_{\emp*\rho}(T) \rightarrow \mathcal{M}^c_\rho(F)$.
\end{proof} 

Evaluating the Monge gap $\mathcal{M}^c_{\emp*\rho}(T)$ requires solving an OT problem. To alleviate computational issues, we use an entropic regularization $\varepsilon \geq 0$, as introduced in Eq.~\eqref{eq:entropicOT}:
\begin{align}
\label{eq:monge-gap-estimator}
\!\!\mathcal{M}^c_{\emp*\rho, \varepsilon}(T) \eqdef\! \tfrac{1}{n}\! \sum_{i=1}^n c(\*x_i, T(\*x_i)) -\! W_{c, \varepsilon}(\emp*\rho, T \sharp \emp*\rho)\,.
\end{align}

The estimator in Eq.~\eqref{eq:monge-gap-estimator}, while being far more effective to compute, retains many of the appealing properties of the unregularized Monge gap:
\begin{itemize}[leftmargin=.3cm,itemsep=.0cm,topsep=0cm]
\item Choosing $\varepsilon = 0$, one recovers $\mathcal{M}^c_{\emp*\rho, 0}(T) = \mathcal{M}^c_{\emp*\rho}(T)$. 
\item For $\varepsilon > 0$, one has $\mathcal{M}^c_{\emp*\rho, \varepsilon}(T) >0$ (see Appendix~\ref{sec:positivity}). 
\end{itemize}

When we add an entropic regularization, we no longer have $\mathcal{M}_{\emp*\rho, \varepsilon}(T) = 0$ when $T$ is optimal, however $\mathcal{M}_{\emp*\rho, \varepsilon}(T) \simeq 0$, provided that $\varepsilon$ is small enough. 

\subsection{Relation to Cyclical Monotonicity.}\label{subsec:properties}
To gain intuition about what $\mathcal{M}^c_{\emp*\rho}$ quantifies, we introduce the notion of cyclical monotonicity. Recall that a set $\Gamma \subset \Omega \times \Omega$ is $c$-CM if for any $n \in \mathbb{N}$, any set $\{\*x_1, ..., \*x_n\} \times \{ \*y_1, ..., \*y_n\} \subset \Gamma$ and permutation $\sigma \in \mathcal{S}_n$ one has:
\begin{align*}
\sum_{i=1}^n c(\*x_i, \*y_i) \leq \sum_{i=1}^n c(\*x_i, \*y_{\sigma(i)})\,.
\end{align*}
Setting $\*y_i:=T(\*x_i)$, the Monge gap estimator using permutations~\citep[Proposition 2.1]{Peyre2019computational} is:
\begin{equation*}
% \label{eq:birkhoff}
\mathcal{M}^c_{\emp*\rho}(T) = \frac{1}{n}\sum_{i=1}^n c(\*x_i, T(\*x_i)) -  \min_{\sigma \in \mathcal{S}_n} \frac{1}{n}\sum_{i=1}^n c(\*x_i, T(\*x_{\sigma (i)})\,,
\end{equation*}
can therefore be interpreted as a quantification of the violation of the cyclical monotonicity of the set $\Gamma \eqdef \supp \left( (\mathrm{Id}, T) \sharp \rho \right)$, measured on sampled points $\{(\*x_1,T(\*x_1)), ...,  ..., (\*x_n,T(\*x_n))\} \subset \Gamma$. Under the assumptions made on $c$ and $\Omega$, the cyclical monotonicity of that set is equivalent to the optimality of $T$, see \citep[Theorem 1.38, Theorem 1.49]{santambrogio2015optimal}.

\subsection{Properties of the Monge Gap.} 
When the Monge gap w.r.t. $\rho$ of a map $T$ is zero, then it will be also be zero on \textit{any} measure whose support is contained in that of $\rho$. This is a crucial property of our regularizer and a natural extension of \citep{Brenier1987}'s result for the $\ell_2^2$ cost, which states that a map is optimal between $\rho$ onto $T\sharp \rho$, if and only if it is the gradient of a convex potential; assuming that is true, that map will therefore move optimally \textit{any} measure whose support is contained in that of $\rho$.

\begin{proposition}\label{prop:more-do-less} Let $\mu,\nu\in\mathcal{P}(\Omega)$ such that $\supp(\mu)\subset\supp(\rho)$, and a map $T$ s.t. $T\sharp \mu = \nu$.
Then $\mathcal{M}_\rho^c(T) = 0$ implies that $T$ is a $c$-OT map between $\mu$ and $\nu$. 
\end{proposition}
\begin{proof}
Let $T, \mu, \nu$ as above and suppose that $\mathcal{M}_\rho^c(T)=0$. Then, $(\mathrm{Id}, T)\sharp\rho$ is an optimal coupling between $\rho$ and $T \sharp \rho$. Since the cost $c$ is continuous, $\supp\left( (\mathrm{Id}, T)\sharp\rho \right)$ is a $c$-cyclically monotone ($c$-CM) set by virtue of \citep[Theorem 1.38]{santambrogio2015optimal}. Because $\supp(\mu) \subset \supp(\rho)$, one has $\supp\left( (\mathrm{Id}, T)\sharp\mu \right) \subset \supp\left( (\mathrm{Id}, F)\sharp\rho \right)$. Since the $c$-CM property is defined for sets, one has that $\supp\left( (\mathrm{Id}, T)\sharp\mu \right)$ is also $c$-CM. Moreover, since $\Omega$ is compact, $c$ is uniformly continuous and bounded. Hence, cyclical monotonicity of its support implies that the coupling $(\mathrm{Id}, T)\sharp\mu$ is optimal between its marginals thanks to \citep[Theorem 1.49]{santambrogio2015optimal}. Therefore, $T$ is a $c$-OT map from $\mu$ to $\nu$. 
\end{proof}
\paragraph{The Quadratic Case.} We focus now on the Monge gap when $c(\cdot, \cdot) = \|\cdot - \cdot \|_2^2$, abbreviated as $\mathcal{M}_\rho^2$, and study the convexity of both $\mathcal{M}_\rho^2$ and $\mathcal{M}_{\emp*\rho, \varepsilon}^2$ for $\varepsilon \geq 0$.

\begin{proposition}[Convexity of (entropic) empirical Monge Gap]
\label{prop:empirical-convexity}
Let $\varepsilon \geq 0$ and an empirical probability measure $\emp*\rho = \frac{1}{n} \sum_{i=1}^n \delta_{\*x_i}$. Then, $\mathcal{M}_{\emp*\rho, \varepsilon}^2$ is convex on vector fields.
\end{proposition}
\begin{proof}
$\mathcal{M}^2_{\emp*\rho, \varepsilon}(T)$ only depends on $T$ via its values on the support of $\emp*\rho$, namely $\*x_1, ..., \*x_n$. Therefore, we write $\*t_i \eqdef T(\*x_i)$ and study the convexity of:
$$
r(\*T) :=\tfrac{1}{n} \|\*X - \*T\|_F^2 - W_{\ell_2^2, \varepsilon}(\emp*\rho, \rho_{\*T})\,,
$$
where $\*X, \*T\in \mathbb{R}^{n\times d}$ contain observations $\*x_i$ and $\*t_i$ respectively, stored as rows, and $\rho_{\*T}$ is the discrete measure supported on the $\*t_i$. Expanding the squares yields:
\begin{align}
\label{eq:empirical-monge-gas-as-max}
    r(\*T) = \max_{\*P \in U_n} 2 \langle \*T, ( \*P - \tfrac{1}{n}I_n )^\top \*X \rangle + \varepsilon H(\*P)\,,
\end{align}
which proves convexity in $\*T$, as a maximum of linear functions in $\*T$, and therefore in $T$. 
\end{proof}
\begin{corollary}[Convexity of Monge Gap]
\label{cor:convexity}
For any $\rho \in \mathcal{P}(\Omega)$, $T \mapsto \mathcal{M}_{\rho}^2(T)$ is convex on continuous vector fields.
\end{corollary}
\begin{proof}
As convexity is preserved under pointwise convergence, it follows from Proposition \ref{prop:empirical-convexity} with $\varepsilon = 0$ and Lemma \ref{lem:consistency}, which applies since we restrict the domain to continuous vector fields.
\end{proof}

\section{Learning with the Monge Gap}\label{subsec:gen}

\begin{figure*}
         \centering
         \includegraphics[width=2.1\columnwidth]{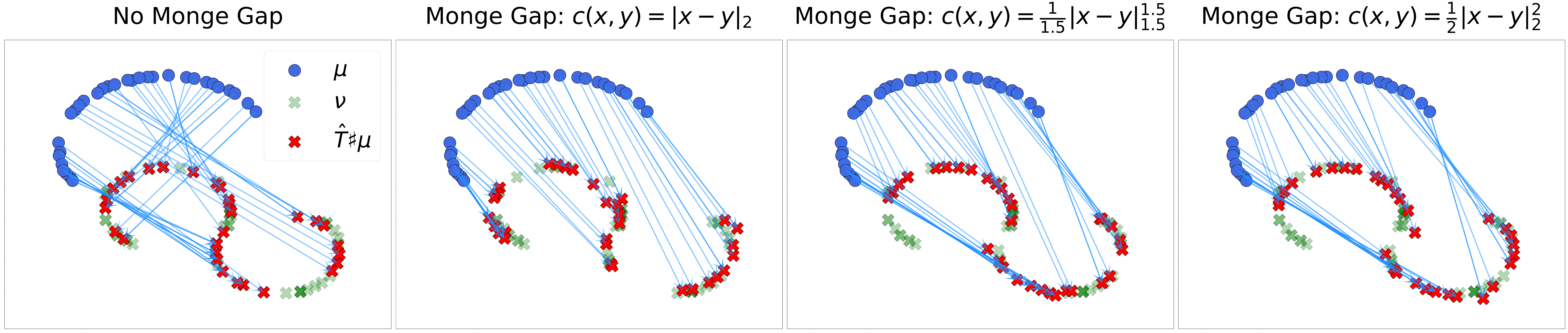}
         \label{fig:lpq_costs}
         \caption{Fitting of transport maps between synthetic measures $\mu$, $\nu$ in dimension $d=2$, with the same fitting loss $\Delta = W_{2, \varepsilon}$ but Monge gap  $\mathcal{M}_\mu^c$ instantiated with various costs $c$. We also fit an MLP without Monge gap, minimizing only the fitting loss. For $c(\*x, \*y) = \|\*x - \*y\|_2$, we use the method for generic costs \S \ref{subsec:gen}, directly parameterizing $T_\theta$ as an MLP and using $\lambda_{\mathrm{MG}} = 5$. For $c(\*x, \*y) = \tfrac{1}{1.5}\|\*x - \*y\|_{1.5}^{1. 5}$ and $c(\*x, \*y) = \tfrac{1}{2}\|\*x - \*y\|_2^2$, since they have the form $c(\*x, \*y) = h(\*x - \*y)$ with $h$ strictly convex and kwown Legendre transform $h^*$, we use the method for costs with structure \S \ref{subsec:struc}. Accordingly, we parameterize $T_\theta = \Id - \nabla h^* \circ F_\theta$ with an MLP $F_\theta$ and penalize lack of conservativity with $\mathcal{C}_\mu$. Moreover, we use $\lambda_{\mathrm{MG}} = 1$ and $\lambda_{\mathrm{cons}} = 0.01$. 
         }
         \label{dimension_plot}
\end{figure*}

We show how the Monge gap can be used to learn approximately $c$-optimal parameterized maps, for any $c$. 

\subsection{Using directly the Monge gap as a regularizer.}\label{subsec:gen}
Let $\mu$, $\nu \in \mathcal{P}(\Omega)$ the source and target measures, and a parameterized family of maps $\{T_\theta\}_{\theta \in \mathbb{R}^p}$. The OT problem~\eqref{eq:monge-problem} balances two goals: (i) ensure $T_\theta \sharp \mu \approx \nu$, while (ii) minimizing the averaged $c$-cost of this displacement. The Monge gap will handle (ii) elegantly, through a convex and non-negative regularization. To handle (i), any fitting loss defined through a divergence $\Delta$ would work. Introducing a regularization weight $\lambda_\mathrm{MG} \geq 0$, this translates to:
\begin{align}
\label{eq:optimization-problem-monge-gap-only}
\min_{\theta\in\R^p} \mathcal{L}(\theta):=\underbrace{\Delta(T_\theta \sharp \mu, \nu)}_{\mathrm{fitting}} + \underbrace{\lambda_\textrm{MG} \, \mathcal{M}_\rho^c(T_\theta)}_{\mathrm{c-optimality}}\,.
\end{align}

\paragraph{On the choice of $\lambda_\mathrm{MG}$.} Assume that a $c$-OT map between $\mu$ and $\nu$ exists, $\Delta$ is a distance, and $\supp(\mu) \subset \supp(\rho)$. Omitting the parameterization, let us consider:
\begin{align*}
\min_{T : \mathbb{R}^d \rightarrow \mathbb{R}^d} \tilde{\mathcal{L}}(T):=\Delta(T \sharp \mu, \nu) +\lambda_\textrm{MG} \, \mathcal{M}_\rho^c(T)\,.
\end{align*}
For any $\lambda_\mathrm{MG} > 0$, the above argmin set matches exactly the $c$-OT maps. Indeed, $\tilde{\mathcal{L}}(T) \geq 0$ with equality i.f.f.\ $\Delta(T\sharp\mu, \nu) = 0$ and $\mathcal{M}^c_\rho(T) = 0$, i.e.\ $T\sharp\mu = \nu$ and $T$ is optimal between $\mu$ and $T_\sharp \mu = \nu$ using Proposition \ref{prop:more-do-less} because $\supp(\mu) \subset \supp(\rho)$. On the contrary, considering naively $R(T) = \int c(\*x, T(\*x)) \diff \mu(\*x)$ as regularizer, one recovers $T^\star$ only if $\lambda_R \rightarrow +\infty$, which highlights why we subtract the optimal transport cost in $\mathcal{M}^c_\rho$. When using parameterized map $T_\theta$, one can simply choose $\lambda_\mathrm{MG}$ to balance the two terms of Optimization Problem \eqref{eq:optimization-problem-monge-gap-only} objective function. When it makes sense, one can set $\Delta = W_c$ and $\lambda_\mathrm{MG} = 1$, so that $\Delta$ and $\lambda_\mathrm{MG}\mathcal{M}_\rho^c$ are naturally homogeneous. 

\paragraph{Gradient of Monge Gap.} Assume from now that $c$ and $\Delta$ are differentiable and let $\epsilon >0$. Optimization Problem \eqref{eq:optimization-problem-monge-gap-only} can be solved by sampling batches $\emp*\mu, \emp*\nu, \emp*\rho$, and considering stochastic gradients.
To better understand the effect of adding the Monge gap to the fitting loss, we take a closer look at the gradient of the Monge gap, $\nabla_\theta \mathcal{M}_{\emp*\rho, \epsilon}^c(T_\theta)$. Since $\varepsilon > 0$ the optimal transport plan $\*P^\varepsilon$ between $\emp*\rho$ and $T_\theta \sharp \emp*\rho$ is unique. Afterwards, thanks to the \citet{danskin2012theory} Theorems, $\mathcal{M}_{\emp*\rho, \epsilon}^c$ is differentiable and its gradient reads:
\begin{equation*}
\nabla_\theta \mathcal{M}^c_{\emp*\rho, \varepsilon} (T_\theta) = \sum_{i,j=1}^n  \left( \tfrac{1}{n} \delta_{ij} - \*P^\varepsilon_{ij} \right) \nabla_\theta c(\*x_i, T_\theta(\*x_j))
\end{equation*}
One can notice that the magnitude of the gradient increases as $\*P^\varepsilon$ deviates from the identity coupling $\frac{1}{n}I_n$ which sends each $\*x_i$ to $T_\theta(\*x_i)$. More precisely, since $\*P^\varepsilon \in U_n$, $\forall i,j, \ 0 \leq \*P^{\varepsilon}_{ij} \leq 1/n$, so:
\begin{equation*}
\left\{
\begin{array}{rcr}
(1 / n) \delta_{ij} - \*P^{\varepsilon}_{ij} \geq 0 \quad \text{if} \quad i=j \quad  \\
(1 / n) \delta_{ij} - \*P^{\varepsilon}_{ij} \leq 0 \quad \text{if} \quad i\ne j \quad 
\end{array}
\right.
\end{equation*}
Using gradient steps on $\theta$ will therefore drive the $T_\theta(\*x_i)$ to make $\*P^{\varepsilon}$ as close as possible to the identity coupling by: decreasing the cost on the diagonal $c(\*x_i,T_\theta(\*x_i))$ while increasing the cost off the diagonal $c(\*x_i,T_\theta(\*x_j))$, $i\ne j$. An experiment showing this dynamic on synthetic data in dimension $d=2$ is provided in Appendix \ref{sec:gradient-flow}. 
\subsection{Handling Costs with Structure.}\label{subsec:struc}
% \label{sec:costs-strcuturing}
The method described in \S \ref{subsec:gen} can be refined when the cost introduces structure in the optimal map. For costs $c(\*x, \*y) = h(\*x - \*y)$ with $h$ striclty convex, the map has structure, as a known functional depending on $h^*$ applied to the gradient a dual potential (see Eq. \eqref{eq:optimal-map}). Accordingly, we can adapt the map's parameterization, introducing a parametrized vector field $F_\theta$ to model directly the dual potential gradient $\nabla \varphi^\star$:
\begin{align}
\label{eq:map_structure_ti_cost}
T_\theta : \*x \mapsto \*x - \nabla h^* \circ F_\theta(\*x).
\end{align}
This case includes notably all $h = \frac{1}{p} \|\cdot\|_p^p$ with $p \geq 1$ and $q$ s.t.\ $\frac{1}{p} + \frac{1}{q} = 1$, for which $h^* = \frac{1}{q} \|\cdot\|_q^q$ 

\textbf{Penalizing Lack of Conservativity.} 
Since $F_\theta$ intends to parameterize a conservative vector field, we follow recent papers that propose a regularization penalizing a lack of conservativity~\cite{chao2022quasi}. By virtue of the Poincaré's lemma \citep[Theorem 4.1, Chap. V]{lang2001fundamental}, on a star shaped domain $S \subset \mathbb{R}^d$, any closed differential form is exact, namely any differentiable vector field whose jacobian is symmetric on $S$ is a gradient field. Introducing a reference measure $\rho$ and considering a differentiable vector field $F$, the regularizer hence penalizes the asymmetry of $\jac_\*x F$ for $\*x \sim \rho$:
\begin{align}
\mathcal{C}_\rho(F) =\mathbb{E}_{X \sim \rho} \left[\|\jac_X F - \jac_X^T F \|_{2}^2\right]\,.
\end{align}
The regularizer $F\mapsto \mathcal{C}_\rho(F)$ is convex on differentiable vector fields. Indeed, for any $\*x$, $F \mapsto \|\jac_\*x F - \jac_\*x F^\top\|_2^2$ is convex as the composition of a linear operator and a convex function, so the convexity of $\mathcal{C}_\rho$ follows from linearity of the expectation. Similar to the Monge gap, we use an empirical estimator for $\mathcal{C}_\rho(F)$. However, for large dimension $d$, computing the full Jacobian $\jac_{\*x} F$ might be too costly. We use instead the \citet{hutchinson1990stochastic} trace estimator, which turns pointwise  Jacobians to pointwise Jacobian vector products (JVPs) and vector Jacobian products (VJPs):
\begin{align*}
\mathcal{C}_\rho(F) =\mathbb{E}_{(X, V) \sim \rho \otimes \mathcal{N}(0, \Id)} \left[\|\jac_X F V - \jac_X^T F V \|_{2}^2\right]
\end{align*}
whose empirical counterpart translates to
\begin{align*}
\mathcal{C}_{\emp*\rho}(F) = \sum_{i=1}^n \sum_{j=1}^m \|\jac_{\*x_i} F \*v_j - \jac_{\*x_i} F^\top \*v_j \|_2^2\,,
\end{align*}
with $\*x_1, ..., \*x_n \sim \rho$ and $\*v_1, ..., \*v_m \sim \mathcal{N}(0, \Id)$. 
Using the JAX framework \cite{jax2018github}, these operations can be carried using the \texttt{jax.vjp} and \texttt{jax.jvp} operators.
\begin{figure*}
         \centering
         \includegraphics[width=2.05\columnwidth]{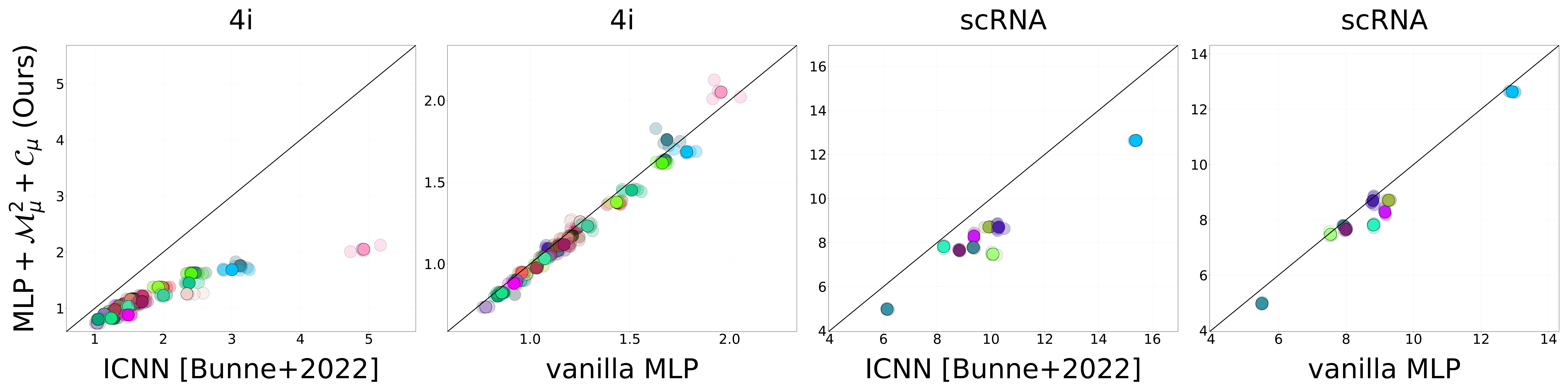}
         \label{fig:diagonal-plot-single-cell}
         \caption{Fitting of a transport map $\hat{T}$ to predict the responses of cells populations to cancer treatments, on 4i and scRNA datasets, providing respectively 34 and 9 treatment responses. For each profiling technology and each treatment, we compare the predictions of a MLP trained with Monge gap $\mathcal{M}_\mu^2(F)$ + conservative regularizer $\mathcal{C}_\mu$ to those provided by a vanilla MLP (trained without regularization), and a gradient-ICNN learned via the neural dual formulation~\cite{makkuva2020optimal}. We measure predictive performance using the Sinkhorn divergence between a batch of unseen (test) treated cells and a batch of unseen control cells mapped with $\hat{T}$, see\S~\ref{sec:single-cell-genomics} and Appendix \ref{sec:numerical_details_single_cell_genomics} for details. Each scatter plot displays points $z_i = (x_i, y_i)$ where $y_i$ is the divergence obtained by our method and $x_i$ that of the other baseline, on all treatments. A point below the diagonal $y=x$ refers to an experiment in which our methods outperforms the baseline. To each treatment, we assign a color and plot 5 runs, along with their mean (the brighter point).}
         \label{dimension_plot}
\end{figure*}
\begin{align}
\label{eq:optimization-problem-monge-gap-conservative-regularizer}
\begin{split}
\min_{\theta\in\R^p} & \mathcal{L}(\theta) 
 := \underbrace{\Delta((\Id - \nabla h^* \circ F_\theta) \sharp \mu, \nu)}_{\mathrm{fitting}} 
\\ & + \underbrace{\lambda_\textrm{MG} \, \mathcal{M}_\rho^c(\Id - \nabla h^* \circ F_\theta)}_{\mathrm{c-optimality}} 
 + \underbrace{\lambda_\textrm{cons} \, \mathcal{C}_\rho(F_\theta)}_{\mathrm{conservativity}}
\end{split}
\end{align}
Note that the Monge gap and the conservative regularizer are not applied to the same vector field. While $\mathcal{M}^c_\rho$ is applied to $T_\theta := \Id - \nabla h^* \circ F_\theta$, $\mathcal{C}_\rho$ is evaluated on $ F_\theta$, to mimic the gradient of a dual potential $\nabla \varphi^\star$. Since $\varphi^\star$ can always be taken $c$-concave \citep[Remark 1.13]{santambrogio2015optimal}, $F_\theta$ can be thought as a \textit{soft} Input $c$-concave Gradient Network.

\section{Related works}
\label{sec:related-work}

\paragraph{Neural OT map estimation.} As recalled in the introduction, duality theory can guide the choice of neural OT architectures, using $c$-concavity. This motivates naturally ICNNs for squared-Euclidean costs, but also more general $c$-concave neural potentials. These approaches are, however, fairly difficult to train and parameterize in practice. \citet{fan2020scalable} propose an alternative approach, conceptually similar to a Wasserstein GAN~\citep{arjovsky2017wasserstein}, where a Lagrange multiplier $f$ is introduced in the Monge formulation defined in Eq.~\eqref{eq:monge-problem} to account for the push-forward constraint $T\sharp \mu=\nu$. This results in a saddle point problem $\sup_f \inf_T \mathcal{L}(f, T)$, trading off two terms, a displacement cost and a fitting loss error. The goal is then to make that displacement cost small, while reaching a fitting loss as close as possible to zero. The proper trade-off between the two terms is, however, difficult to get right: the displacement cost cannot be minimized to zero (that term represents the ``travelled'' distance to go from source to target), and its scale will interfere with that the fitting loss (which should be, ideally, close to 0). By contrast, in our approach both the fitting loss and the Monge gap (which can be interpreted as a ``recentered'' displacement cost) should be close to $0$. In that sense the Monge gap is truly a regularizer, and not a displacement cost.

\paragraph{Beyond maps.} Similar to the approach taken with the Monge formulation, the Kantorovitch formulation can also be reformulated as a saddle point problem, by relaxing $\pi \in \Pi(\mu, \nu)$ to $\pi \in \Pi(\mu)$ and introducing a Lagrange multiplier for the second marginal constraint. A recent line of work proposes to directly estimate non deterministic parameterized couplings $\pi_\theta \in  \Pi(\mu)$, modelling $\pi_\theta(\*y|\*x)$ via "one to many" stochastic maps \cite{korotin2022kernel, korotin2022neural, korotin2022general, gazdieva2022unpaired}. More precisely, for a latent space $\mathcal{Z}$, take $\gamma \in \mathcal{P}(\mathcal{Z})$ and a stochastic map $T_{\pi_\theta} : \Omega \times \mathcal{Z} \rightarrow \Omega$, if $\*x \sim \mu$, then for any  $\*z \sim \gamma$, $(\*x, T_{\pi_\theta}(\*x, \*z)) \sim \pi_\theta \in \Pi(\mu)$. Imposing deterministic couplings $\pi_\theta = (\Id, T_\theta)\sharp\mu$, we recover the saddle point Monge problem of \citet{fan2020scalable}, which is why we only consider \citet{fan2020scalable} in our experiments. %Adding stochasticity in the coupling $\pi_\theta$ by diffusing the conditional law $\pi_\theta(\*y|\*x)$ seems to stabilize performances and improve the generatie power. However, the fitting of $\pi_\theta$ still requires the resolution of a difficult saddle point problem. 
%An exciting direction for future work would be to naturally extend the Monge Gap to define the Kantorovich gap measuring the sub-optimality of $\pi_\theta$, allowing regularized and \textit{non max-min} neural coupling fitting.

\section{Experiments}
\label{sec:experiments}

\begin{figure}[h]
         %\centering
         \hspace{-90mm}
         \label{fig:spherical-fitted-map-arccos}
         \includegraphics[width=3.1\columnwidth]{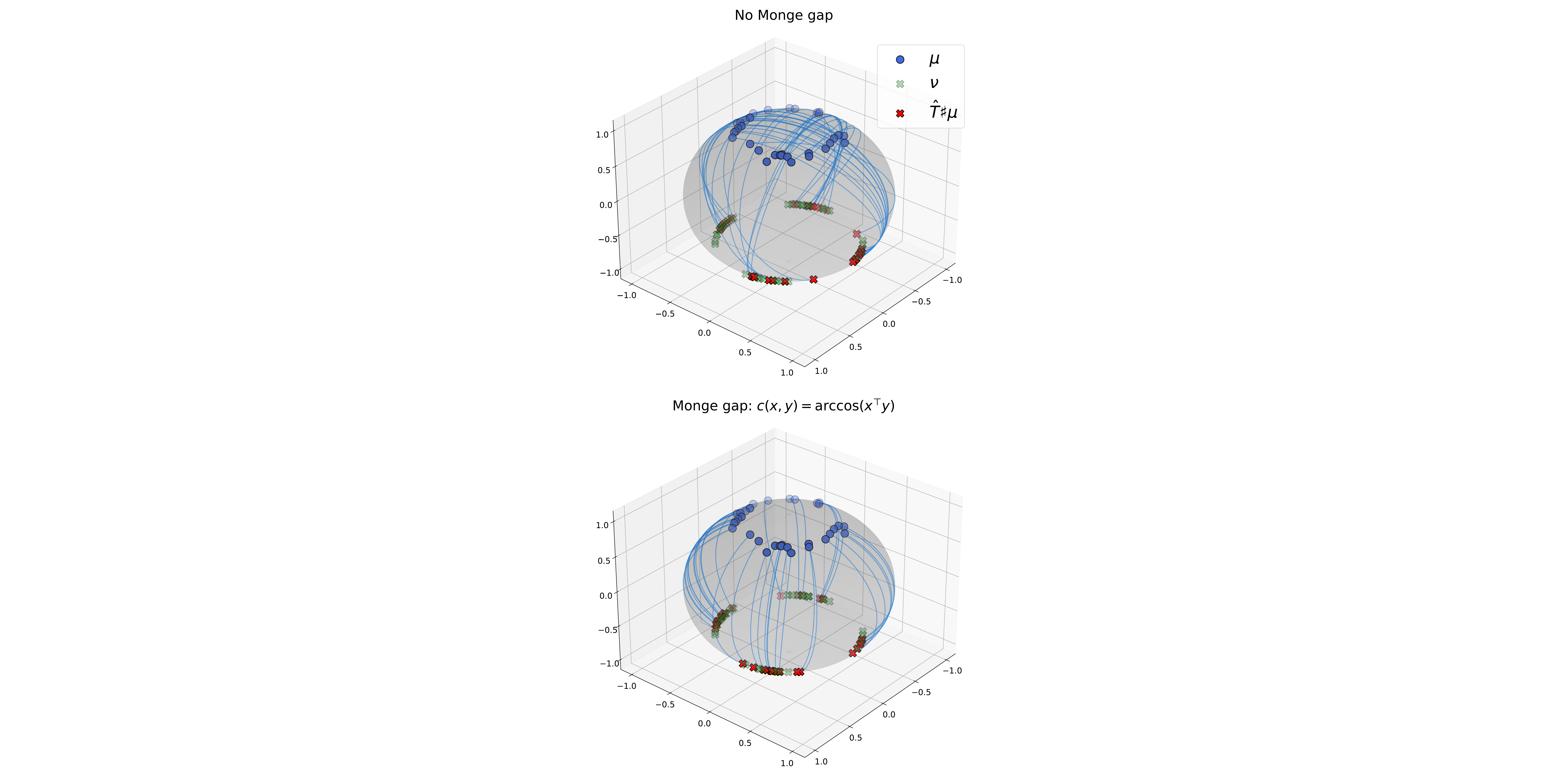}
         \caption{Fitting of transport maps betwen synthetic measures on the $2$-sphere. In both cases, we parameterize the map as $T_\theta = F_\theta / \|F_\theta\|_2$ where $F_\theta$ is an MLP, and we use $\Delta = W_{\ell_2^2, \varepsilon}$ as fitting loss. On the upper plot, we do not use any regularizer while on the lower plot we regularize with the Monge gap instantiated for the geodesic cost $c(\*x, \*y) = \arccos(\*x^\top\*y)$ and use $\lambda_\mathrm{MG} = 1$.}
\end{figure}

\begin{figure}
         \centering
         \includegraphics[width=1\columnwidth]{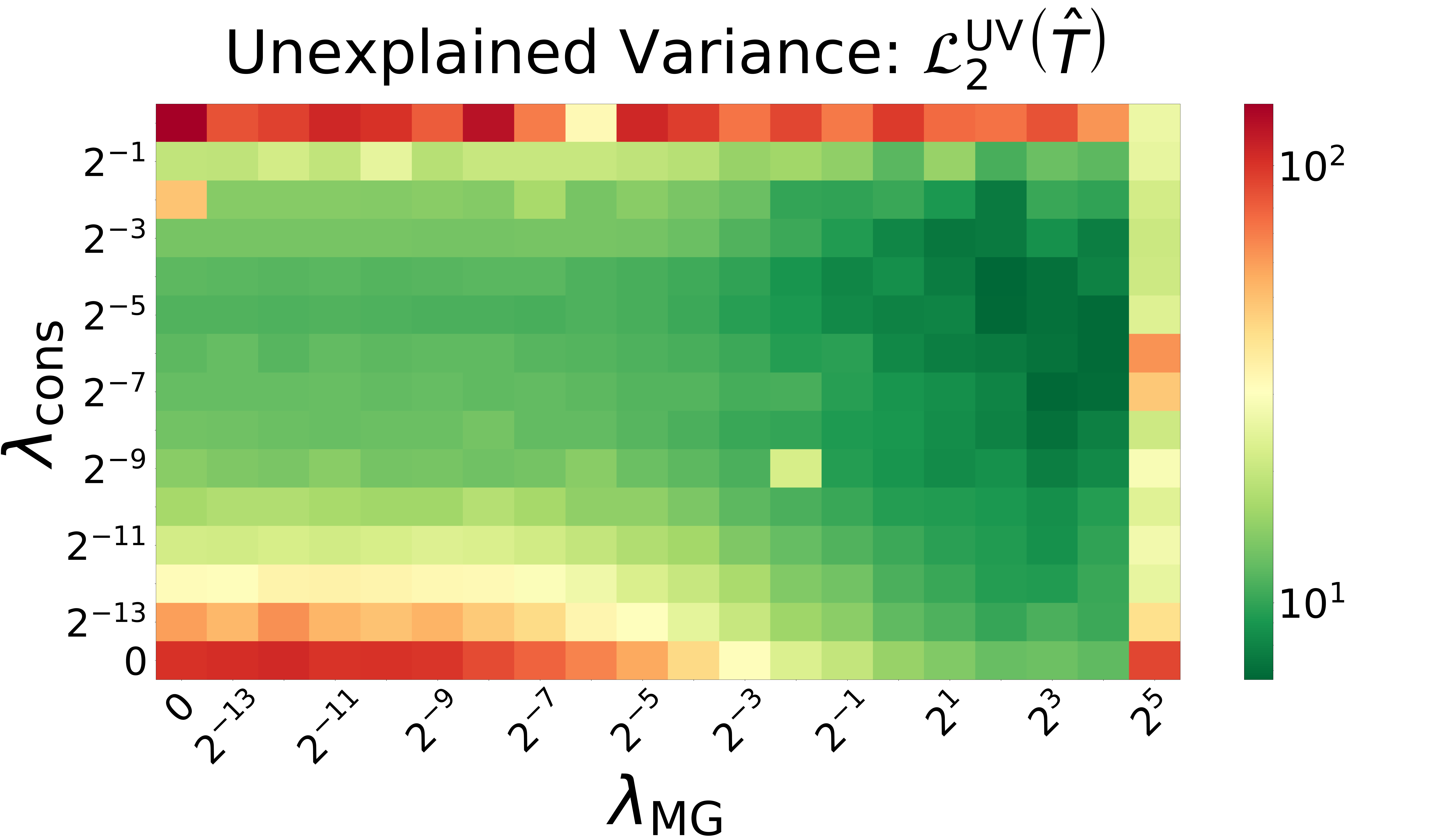}
         \label{fig:heatmap-unexplained-variance}
         \caption{
         Heatmap showing the influence of the Monge gap $\mathcal{M}_\mu^2$ and the conservative regularizer $\mathcal{C}_\mu^2$, when learning the Monge map for the $\ell_2^2$ cost between \citet{korotin2021do} benchmark pair of dimension $d=32$. For each pair of regularization weights $(\lambda_\mathrm{MG}, \lambda_\mathrm{cons})$ on a regular grid.
         we report the unexplained variance $\mathcal{L}_2^{\mathrm{UV}}(\hat{T})$ provided by the the estimated map $\hat{T}$. 
         }
\end{figure}

\begin{figure}
         \centering
         \label{fig:all_dimensions_plot}
         \includegraphics[width=0.95\columnwidth]{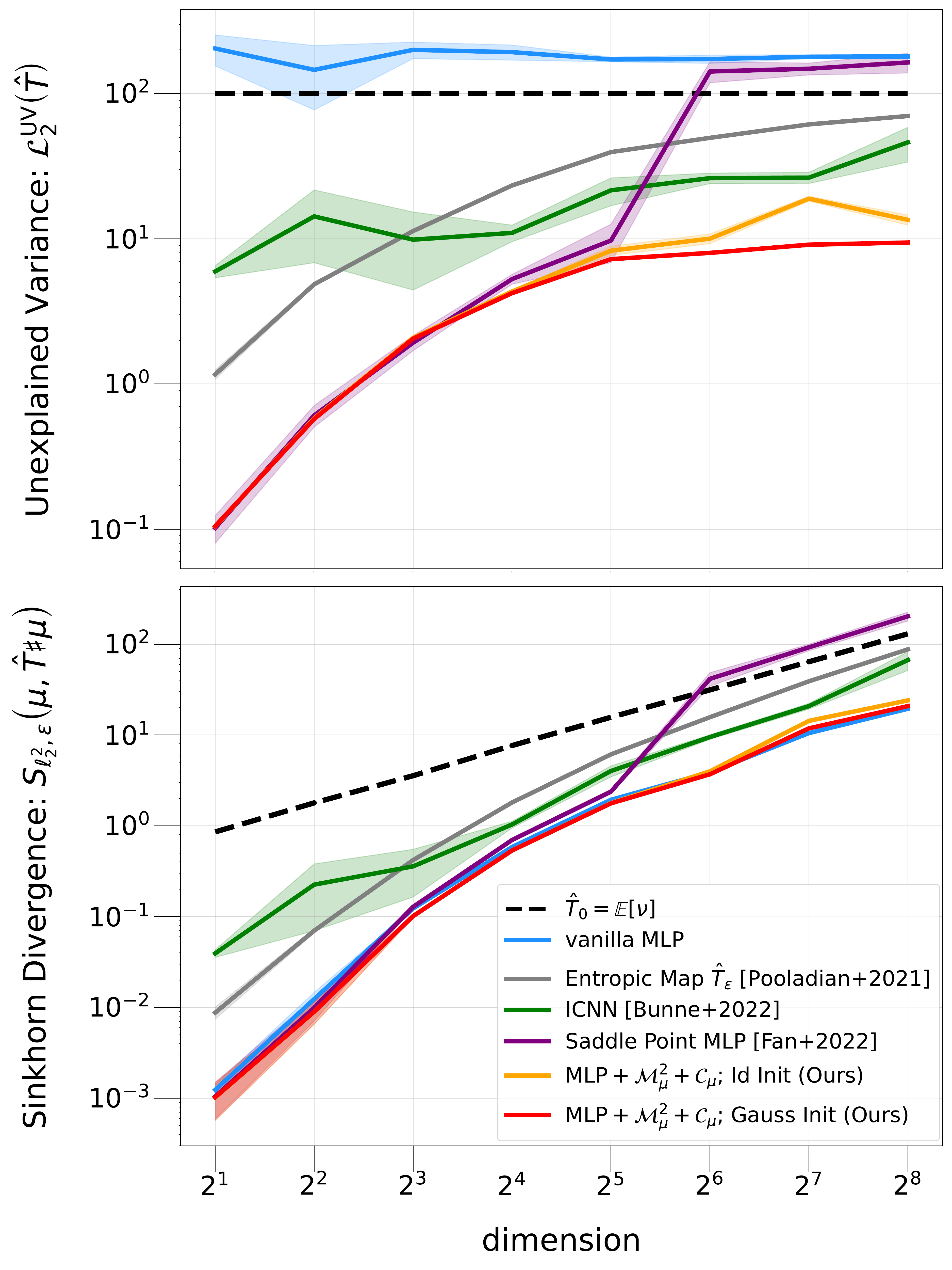}
         \caption{Performances of Monge gap-based learning and baselines on estimating the ground-truth maps between each pair of Gaussian mixtures $\mu$, $\nu$ in dimension $d \in \{2, 4, 8, ..., 256\}$ of the \citet{korotin2021do} benchmark. We report both Sinkhorn divergence $S_{\ell_2^2, \varepsilon}(\hat{T}\sharp \mu, \nu)$ and the unexplained variance $\mathcal{L}_2^{\mathrm{UV}}(\hat{T})$ averaged over 5 fittings.}
         \label{fig:dimension_plot}
\end{figure}

We evaluate the ability of our method to recover OT maps between both synthetic (\S\ref{sec:synthetic-data},\ref{sec:korotin-benchmark}) and real (\S\ref{sec:single-cell-genomics}) datasets.

\subsection{Experimental Setting.} 
\textbf{Reference measure.} Choosing the reference measure $\rho$ is the first step in our construction. We provide a simple and preliminary toy experiment in Appendix \ref{sec:additional_experiments}. We settle in practice for the simplest choice of setting $\rho = \mu$ and leave other choices for future research.

\textbf{Transport map fitting.} When $c(\*x, \*y) = h(\*x - \*y)$ with $h$ strictly convex, we use the method provided in section \S \ref{subsec:struc}. In particular, for all experiments carried out with the quadratic cost (\S ~\ref{sec:korotin-benchmark}, \S ~\ref{sec:single-cell-genomics}), we parameterize the map as $T_\theta = \Id - F_\theta$ where $F_\theta$ is an MLP and use both $\mathcal{M}^2_\mu$ and $\mathcal{C}_\mu$ as regularizers. Otherwise, we use the generic cost method (\S \ref{subsec:gen}) by parameterizing $T_\theta$ as an MLP and using only the Monge gap. All MLPs, trained with or without regularizers, are fitted with $\Delta = W_{\ell_2^2, \varepsilon}$. We adapt \citet{bunne2022supervised}, to define both Identity and Gaussian initialization schemes for our neural transport maps, see details in Appendix \ref{sec:initializers}. See Appendix \ref{sec:numerical_details} for details about other hyperparameters.

\textbf{Metrics. } To measure the predictive performances of an estimator $\hat{T}$ of $T^\star$,
 we rely on (i) the Sinkhorn Divergence between the target and the fitted target measures, namely $S_{\ell_2^2, \varepsilon}(\nu, \hat{T} \sharp \mu)$ and, when $T^\star$ is known, (ii) the $\mathcal{L}_2$ unexplained variance percentage \cite{makkuva2020optimal}, \cite{korotin2020wasserstein}, \cite{korotin2021do} defined as: 
\begin{align}
\label{eq:unexplained_variance}
\mathcal{L}_2^{\mathrm{UV}}(\hat{T}) \eqdef 100 \cdot \frac{\mathbb{E}_\mu[\|\hat{T}(X) - T^*(X)\|^2}{\mathrm{Var}_\nu(X)}\,.
\end{align}
$S_{\ell_2^2, \varepsilon}(\nu, \hat{T} \sharp \mu)$ quantifies the generative power of the method, as a valid divergence between the reconstructed and the actual target. For all experiments, we use $\varepsilon = 0.1$. Instead, $\mathcal{L}_2^{\mathrm{UV}}(\hat{T})$ quantifies not only this generative power but the Monge optimality, measuring the deviation of $\hat{T}$ from $T^\star$. This deviation is normalized by the variance of $\nu$, so that the constant baseline $\hat{T}_0 = \mathbb{E}_\nu[Y]$ provides $\mathcal{L}_2^\mathrm{UV}(\hat{T}_0) = 100 \%$. 

\subsection{Synthetic Data.}

\label{sec:synthetic-data}

\textbf{$\ell_p^q$ costs. } We evaluate both \S ~\ref{subsec:gen} and \S ~\ref{subsec:struc} methods on $(\ell_p^q)_{p,q \geq 1}$ costs, see Figure \ref{fig:lpq_costs}. For $c(\*x, \*y) = \|\*x - \*y\|_2$, it highlights the “nocrossing” property of OT maps for costs that are distances, hence satisfying triangular inequality.

\textbf{Costs on the sphere. } We consider measures supported on the $2$-sphere along with the geodesic cost $c(\*x, \*y) = \arccos(\*x^\top \*y)$, see Figure \ref{fig:spherical-fitted-map-arccos}. Additional experiments on the $2$-sphere can be found in Appendix \ref{sec:additional_experiments}.

\subsection{High Dimensional Benchmark Pairs.}
\label{sec:korotin-benchmark}

To assess that our method allows to recover Monge maps, we use the \citet{korotin2021do} benchmark, providing pairs of Gaussians mixtures $\mu$, $\nu$ in dimension $d \in \{2, 4, 8, ..., 256\}$, for which the optimal map for the squared Euclidean cost is known, as the gradient of a sum of two ICNNs $\psi_1, \psi_2$. 

\textbf{Baselines. } We compare our method to: (i) a vanilla MLP fitted without regularization, (ii) the ICNN neural dual formulation with Gaussian initializer \cite{bunne2022supervised} (iii) an MLP trained via the saddle point problem \cite{fan2020scalable}, (iv) the entropic map \cite{pooladian2021entropic} and (v) the constant map $\hat{T}_0 = \mathbb{E}_\nu[Y]$. Note that we use the ICNN architecture provided by \citep[Section 4]{bunne2022supervised}, which is not the one used for $\psi_1, \psi_2$. This slightly mitigates the bias favoring ICNN-based methods induced by the benchmark pair design.

\textbf{Effects of Hyperparameters. } 
We first assess the impact of regularization weights $(\lambda_\mathrm{MG}, \lambda_\mathrm{cons})$ on the estimation. We fit a map between the \citet{korotin2021do} benchmark pair when $d = 32$ and report the unexplained variance $\mathcal{L}^{\mathrm{UV}}_2(\hat{T})$ by varying the weights on a regular grid. The results are shown on Figure \ref{fig:heatmap-unexplained-variance}. For small regularizers, we learn an arbitrary pushforward, leading to poor performance. As both regularizations increase, especially $\lambda_\mathrm{MG}$, $\hat{T}$ gets closer to $T^\star$. Interestingly, we observe that the region of the heatmap for which $(\lambda_{\mathrm{MG}}, \lambda_{\mathrm{cons}})$ provides good performances is wide, showing robustness to hyperparameter choice. Following these results, we use $(\lambda_{\mathrm{MG}}, \lambda_{\mathrm{cons}}) = (1, 0.01)$ for $d \leq 64$ and $(\lambda_{\mathrm{MG}}, \lambda_{\mathrm{cons}}) = (10, 0.1)$ for $d \geq 128$ for the experiments on the whole benchmark \ref{fig:all_dimensions_plot}, regularizing slightly more in high dimensions.

\textbf{Results.} See Figure \ref{fig:all_dimensions_plot}. As expected, a vanilla MLP without regularization learns a pushforward that does not generalize as well as approaches trained with regularizers. For low $d \leq 8$, the saddle point estimator \cite{fan2020scalable} remains competitive. Our method, trained along with Gaussian initializer, performs uniformly better than the baselines for $d \geq 16$. This gap widens for $d \geq 64$, when the saddle point estimator starts yielding very poor results, worse than the constant baseline in terms of both generative power and Monge optimality. The ICNNs give unstable and moderate performances, despite the Gaussian initializer scheme, highlighting the difficulty of their training. 

\subsection{Single-Cell Genomics.}
\label{sec:single-cell-genomics}

\textbf{Experimental setting.} Predicting the response of cells to a perturbation is a central question in biology. In this context, feature descriptions of control and treated cells can be treated as probability measures $\mu$ and $\nu$, and perturbation fitted as a transport map $\hat{T}$. Following~\cite{schiebinger2019}, the use of OT theory to recover this map $\hat{T}$ has been used~\cite{bunne2022proximal, bunne2021learning, bunne2022supervised, lubeck2022neural, eyring2022modelling}. We predict responses of cells populations to cancer treatments (perturbations) using the proteomic dataset used in~\citep{bunne2021learning}, consisting of two melanoma cell lines. Patient data is analyzed using (i) 4i \cite{gut2018multiplexed} and scRNA sequencing \cite{tang2009mrna}. For each profiling technology, the response to respectively (i) 34 and (ii) 9 treatments are provided. As in \cite{bunne2021learning}, (i) training is performed with the quadratic cost, in the data space for the 4i data and in a latent space learned by the scGen autoencoder \cite{lotfollahi2019scgen} for the scRNA data and (ii) both evaluations are carried in data space, selecting the top 50 marker genes for scRNA data using the \texttt{scanpy} \cite{wolf2018scanpy} function \texttt{rank\_genes\_groups}. We fix the regularization weights for all treatments of each datatype: $(\lambda_\mathrm{MG}, \lambda_\mathrm{cons}) = (1, 0.01)$ for 4i and $(\lambda_\mathrm{MG}, \lambda_\mathrm{cons}) = (10, 0.1)$ for scRNA.

\textbf{Baselines.} We compare our method to: (i) a vanilla MLP fitted without regularization, (ii) the ICNN neural dual formulation with Gaussian initializer \cite{bunne2022supervised}.

\textbf{Results} are shown in Figure \ref{fig:diagonal-plot-single-cell}. On both 4i or scRNA data, our method gives a better prediction. These results also shows that standard MLPs trained without regularization should not be discarded as a poor contender, since they perform consistently better than ICNNs. Our regularizers $\mathcal{M}_\rho^2$ and $\mathcal{C}_\mu$ improve performance further. We believe this illustrates the rigidity of the ICNN architecture~\cite{korotin2021do, amos2022amortizing}).

\textbf{Conclusion.} We have provided in this paper a novel strategy to train optimal transport maps. Our approach is grounded on regularization rather than on constraints. We provide a regularizer, the Monge gap, that has many favorable properties: lower-bounded by 0, and 0 when the property is observed, with a scale (as a difference between averaged distances) that is comparable to that of a fitting loss. That regularizer allows a more efficient trade-off to train maps that should be OT-like, rather than exactly conforming to OT theory. The regularizer adapts to any cost $c$, but requires defining a reference measure $\rho$. An interesting direction lies in trying to come up with adaptive ways to define that measure, linking it to data measures of interest.

\bibliographystyle{abbrvnat}
\bibliography{main}

\onecolumn
\newpage
\appendix
\section{Proofs}
\label{proofs}

\subsection{On the Positivity of $\mathcal{M}^c_{\emp*\rho, \varepsilon}$}
\label{sec:positivity}
Recall that 
\begin{align*}
\!\!\mathcal{M}^c_{\emp*\rho, \varepsilon}(T) \eqdef\! \tfrac{1}{n}\! \sum_{i=1}^n c(\*x_i, T(\*x_i)) -\! W_{c, \varepsilon}(\emp*\rho, T \sharp \emp*\rho)\,.
\end{align*}
Indeed, for any coupling $\*P \in U_n$, and since $-\varepsilon H(\*P) < 0$, one has:
$$\langle \*P, \*C\rangle -\varepsilon H(\*P) < \langle \*P, \*C\rangle$$
As a result, applying minimization on both sides yields that $W_{c, \varepsilon}(\emp*\rho, F\sharp\emp*\rho) < W_{c, 0}(\emp*\rho, F\sharp\emp*\rho)$, and therefore: 
$$
\mathcal{M}^c_{\emp*\rho, \varepsilon}(F) > \mathcal{M}^c_{\emp*\rho, 0}(F) = \mathcal{M}^c_{\emp*\rho}(F)\geq 0.
$$
\section{Numerical Details}
\label{sec:numerical_details}

\subsection{Initializer Schemes}
\label{sec:initializers}
\begin{figure*}[h]
         \centering
         \includegraphics[width=0.9\columnwidth]{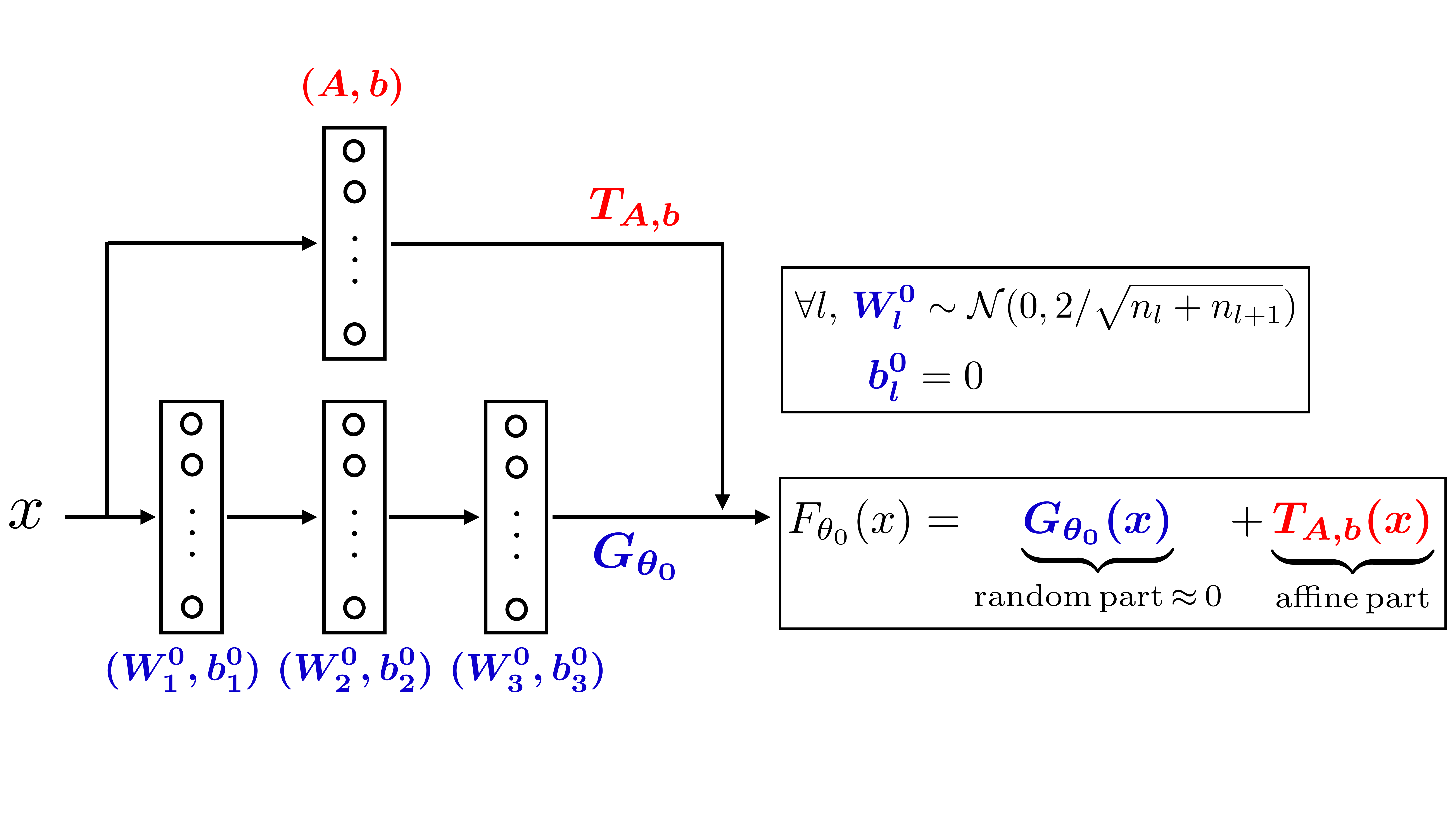}
         \label{fig:affine_init}
         \caption{Initialization scheme to match affine maps applied to a 3 hidden layers MLP. We initialize the feedforward weights using the \citet{Glorot2010UnderstandingTD} initialization technique, and add a residual layer matching the targeted affine map.}
         \label{dimension_plot}
\end{figure*}

Let $F_\theta : \mathbb{R}^d \rightarrow \mathbb{R}^d$, $\theta \in \mathbb{R}^p$, an MLP.
For any affine map $T_{A,b} : \*x \in \mathbb{R}^d \mapsto A\*x +b$ with $A \in \mathbb{R}^{d \times d}, b \in \mathbb{R}^d$, it is simple to choose $\theta_0$ such that $F_{\theta_0} \approx T_{A,b}$. One can initialize the feedforward weights randomly with relatively low variance and add a residual layer from the input layer to the output layer with parameters $(A, b)$. This approach is described in Figure \ref{fig:affine_init}. 

\begin{itemize}[leftmargin=.3cm,itemsep=.0cm,topsep=0cm]
\item \textbf{Identity.} For generic costs, we directly parameterize $T_\theta$ as an MLP, so we initialize with a residual layer parameterizing the identity. For structured costs $c(\*x, \*y) = h(\*x - \*y)$, since we parameterize $T_\theta = \Id - \nabla h^* \circ F_\theta$ with $F_\theta$ an MLP, one typically has that for any $\*x_0 \in \mathbb{R}^d$ close to $0$, $\nabla h^* (\*x_0 ) \approx 0$. Thus, in this case, we don't need to use a residual layer but initializing the feedforward weights randomly with a low variance provides $F_{\theta_0} \approx 0$ so $T_{\theta_0} = \Id - \nabla h^\star \circ F_{\theta_0} \approx \Id$. 
\item \textbf{Gaussian.} This initializer uses the closed form of the OT map between Gaussian measures for the quadratic cost, which is affine. Therefore, it only applies for the quadratic cost, where $T_\theta = \Id - F_\theta$.  We denote $T_{\mathcal{N}}$ the affine OT map between the Gaussian approximations of $\mu$ and $\nu$. First, we estimate $T_{\mathcal{N}}$ from samples, forming empirical means and covariances $(m_{\emp*\mu}, \Sigma_{\emp*\mu})$ and $(m_{\emp*\nu}, \Sigma_{\emp*\nu})$:
\begin{equation}
\hat{T}_{\mathcal{N}} : \*x \mapsto \Sigma_{\emp*\mu}^{-1/2} \left( \Sigma_{\emp*\mu}^{1/2} \Sigma_{\emp*\nu} \Sigma_{\emp*\mu}^{1/2} \right)^{1/2} \Sigma_{\emp*\mu}^{-1/2} (\*x - m_{\emp*\mu}) + m_{\emp*\nu}\,. 
\end{equation}
Square roots and inverse square roots of PSD matrices are computed with the \texttt{OTT-JAX} \cite{cuturi2022optimal} implementation of the \citet{higham_stable_1997} algorithm. Then, we initialize  $F_{\theta_0} \approx \Id - \hat{T}_{\mathcal{N}}$ using a residual layer, hence $T_{\theta_0} \approx \hat{T}_\mathcal{N}$.
\end{itemize}

\subsection{Fixed hyperparameters across experiments.}

\textbf{Entropic regularization.} Whenever we run the Sinkhorn algorithm on a cost matrix $\*C$, we set $\varepsilon = 0.01 \cdot \mathrm{mean}(\*C)$. The only case where we use a different $\varepsilon$ value is for evaluation, when we compute the Sinkhorn divergence $S_{\ell_2^2, \varepsilon}$, for which we set $\varepsilon=0.1$ across all experiments. We use the \texttt{OTT-JAX} \cite{cuturi2022optimal} implementation of the Sinkhorn algorithm.

\textbf{Number of Hutchsinon vectors.} Whenever we use the conservative regularizer, the number of hutchinson vectors $m$ is fixed to the upper integer part of 20\% of the dimension $d$. We remind that the computation of the estimator $\mathcal{C}_{\emp*\rho}(F)$ requires to perform both $ n \cdot m$ JVPs and VJPs.  In order to gain computational efficiency, we obviously need to choose $m \ll d$. Indeed, computing the full Jacobians  $\jac_{\*x_i}F$ and $\jac_{\*x_i}F^\top$ requires the computation of respectively $d$ JVPs and VJPs, instantiated along the vectors of the canonical basis of $\mathbb{R}^d$. 

\textbf{ICNNs.} All ICNNs are trained with the \texttt{NeuralDualSolver} of \texttt{OTT-JAX} which uses the \citet{bunne2022supervised} Gaussian initializer and hence the induced specific architecture. As suggested by \citet{makkuva2020optimal} and used in \citet{bunne2021learning, bunne2022supervised}:
\begin{itemize}[leftmargin=.3cm,itemsep=.0cm,topsep=0cm]
\item To represent discontinuous transport maps, it uses $\mathrm{ReLu}$ as activation function. 
\item It relaxes the positivity constraint on the feedforward weights $W_k^z$ of the ICNN $g_\theta$ s.t.\ $T_\theta = \nabla g_\theta$ with the penalty:
\begin{equation}
    R(\theta) = \sum_{W_k^z \in \theta} \|\max(-W_k^z, 0) \|_F^2
\end{equation}
\end{itemize}

\textbf{MLPs.} All MLPs are vanilla fully connected layers. To train MLPs within the \citet{fan2020scalable} saddle point problem, we follow their choice of using the $\mathrm{PRelu}$ activation function for both the Lagrange multiplier $f$ and the map $T$. For all our MLPs, we use the $\mathrm{GeLu}$ activation \cite{hendrycks2020gelu}.

\paragraph{Calibration of NN sizes.} As the employed ICNN architecture uses (i) linear residual layers from the input layer to each hidden layer and (ii) specific layers designed for Gaussian and identity initializers scheme, if we fix the number of layers and hidden units, they naturally have more parameters than the MLP with same number of layers and hidden units. In particular, the  layers suited to the initializers scheme are quadratic in the input, so the difference in parameters explodes as the dimension increases. For instance, for data in dimension $d=64$, an ICNN with hidden layer sizes [128, 64, 64] has 33,345 parameters, while an MLP with same hidden layer sizes and a residual layer from input layer to output layer for Gaussian initialization (see \S~\ref{sec:initializers}) has 24,896 parameters. Thus, the ICNN has about 33\% more parameters than the MLP. To mitigate this difference, for each experiment where we use both an ICNN and an MLP, we first fix the ICNN size, then we use an MLP with the same number of layers but we adapt the number of hidden units on each of its layers to match the number of parameters up to $1\%$. In the previous example, this leads to an MLP with hidden layer sizes [146, 82, 82] which leads to 33,662 parameters.

\subsection{Synthetic Data}

\paragraph{$\ell_p^q$ costs.} We train all MLPs with $W_{\ell_2^2, \varepsilon}$ as fitting loss. For $c(\*x, \*y) = \|\*x - \*y\|_2$, we parametrize $T_\theta$ as an MLP and use $\lambda_{\mathrm{MG}} = 5$. For $c(\*x, \*y) = \tfrac{1}{1.5}\|\*x - \*y\|_{1.5}^{1. 5}$ and $c(\*x, \*y) = \tfrac{1}{2}\|\*x - \*y\|_2^2$, we parametrize $T_\theta = \Id - \nabla h^* \circ F_\theta$ with an MLP $F_\theta$ and add conservativity regularizer $\mathcal{C}_\mu$. We use $\lambda_{\mathrm{MG}} = 1$ and $\lambda_{\mathrm{cons}} = 0.01$ in both cases. Except for the trained MLP without regularization which is randomly initialized, for all other MLPs we use the identity initializer. All MLPs have hidden layer sizes [128, 64, 64]. They are trained with ADAM \cite{kingma2014adam} for $n_{\mathrm{iters}} = 50,000$ iterations with a learning rate $\eta=0.01$ and a batch size $B = 1024$. 

\paragraph{Costs on the sphere.}  We parameterize the maps with $T_\theta = \frac{F_\theta} {\|F_\theta\|_2}$ where $F_\theta$ is an MLP. We train the MLPs with $W_{\ell_2^2, \varepsilon}$ as fitting loss and set $\lambda_\mathrm{MG} = 1$ for both $c(\*x, \*y) = \arccos(\*x^\top \*y)$ and $c(\*x, \*y) = - \log(\*x^\top \*y)$. All MLPs have hidden layer sizes [128, 64, 64] and are randomly initialized. They are trained with ADAM for $n_{\mathrm{iters}} = 10,000$ iterations with a learning rate $\eta=0.01$ and a batch size $B = 1024$. 

\subsection{Korotin Benchmark}

\paragraph{Evaluation.} We compute both the Sinkhorn divergence $S_{\ell_2^2, \varepsilon}(\hat{T}\sharp\mu, \nu)$ and the unexplained variance $\mathcal{L}_2^{\mathrm{UV}}(\hat{T})$ to evaluate the models on $8,192$ unseen samples from the source and the target measures.

\paragraph{ICNNs.} We initialize the ICNNs using the Gaussian initializer scheme instantiated on $4,096$ samples. We optimize them using ADAM for $_\mathrm{iters} = 100,000$ and $n_\mathrm{inner\_iters} = 10$, with a learning rate $\eta = 10^{-4}$ a batch size $B=1024$. For all experiments, we us ICNNs with hidden layer sizes [max(2$d$, 128), max($d$, 64), max($d$, 64)] where $d$ is the dimension of the data.

\paragraph{Our MLPs.} We initialize the MLPs testing both Gaussian and Identity initializer scheme instantiated on $4,096$ samples. We also test the Identity initializer because it generalizes to generic costs. We train MLPs with $W_{\ell_2^2, \varepsilon}$ as fitting loss. When using regularization, we set $\lambda_\mathrm{MG}=1$ and $\lambda_\mathrm{cons}=0.01$ for $d \leq 64$, and $\lambda_\mathrm{MG}=10$ and  $\lambda_\mathrm{cons}=0.1$ for $d \geq 128$. With or without regularizations, we train the MLPs for $n_{\textrm{iters}} = 100,000$ iterations with a batch size $B = 1024$ and the Adam optimizer. For $d \leq 64$  we use a learning rate $\eta = 0.01$, along with a polynomial schedule of power $p=1.5$ to decrease it to $10^{-5}$. For $d \geq 64$ we change the initial learning rate to $\eta = 0.001$ but keep the same polynomial schedule. When using the Gaussian initializer scheme, we instantiate it on $4,096$ samples. We set the hidden layer sizes size according to the size of the ICNNs.

\paragraph{Saddle Point Problem \citet{fan2020scalable} MLPs.} We train the saddle point problem \cite{fan2020scalable} with two MLPs of hidden layer sizes adapted to the ICNN ones. We optimize them using ADAM for $n_\mathrm{iters} = 100,000$ and $n_\mathrm{inner\_iters} = 10$, with a batch size $B=1024$ and a learning rate $\eta = 10^{-4}$, which is the learning rate mostly used in their experiments. For the dimensions $d \geq 64$, we did not succeed in tuning the learning rate to improve the performance.

\paragraph{Entropic map.} We train the entropic map using $8,192$ from the source and the target measures. 

\subsection{Single Cell Genomics}

\label{sec:numerical_details_single_cell_genomics} 

\paragraph{Evaluation.} For each dataset, we perform a 60\%-40\% train-test split on both conrol and treated cells, and evaluate the models on the 40\% of unseen control and treated cells. We perform such a strong train-test split because the datasets are unbalanced: they contain fewer treated cells than control cells. As we evaluate the performances with $S_{\ell_2^2, \varepsilon}$ which is a distributional metric, we need a number of test samples high enough to make this quantity meaningful. To counteract this unbalancedness, \citet{bunne2021learning} makes a 80\%-20\% train-test split but concatenates the training and treated cells for evaluation. We do not follow this strategy to evaluate the models only on unseen treated cells.

\paragraph{MLPs.} We train all MLPs with  $W_{\ell_2^2, \varepsilon}$ as fitting loss. When using regularization, we set $\lambda_\mathrm{MG}=1$ and $\lambda_\mathrm{cons}=0.01$ for the 4i data, and $\lambda_\mathrm{MG}=10$ and  $\lambda_\mathrm{cons}=0.1$ for the scRNA data. With or without regularizations, we train the MLPs for $n_{\textrm{iters}} = 10,000$ iterations with a batch size $B = 512$ and the ADAM optimizer \cite{kingma2014adam} using a learning rate $\eta = 0.001$, along with a polynomial schedule of power $p=1.5$ to decrease it to $10^{-5}$. When using regularization, we initialize with the Gaussian initailizer scheme trained on half of the training set. We set the hidden layer sizes according to the ones of the ICNNs.

\paragraph{ICNNs.} We use the Gaussian initializer scheme trained on half of the training set. We train the ICNNs using ADAM and learning rate $\eta = 10^{-4}$. \citet{bunne2021learning} optimize the ICNNs on $n_{\mathrm{iters}} = 100,000$ and $n_{\mathrm{inner\_iters}} = 10$, with a batch size $B=256$. On the other hand, since we use a batch size $B=512$ for our models and it is a fundamental hyperparameter whose increase can drastically improve performances, especially in OT based models, we adpat the batch size while keeping the same number of epochs: we train the ICNNs on $n_{\mathrm{iters}} = 50,000$ and $n_{\mathrm{inner\_iters}} = 10$ with $B=512$. We initialize ICNNs with Gaussian initializer \cite{bunne2022supervised} using half of the training set. For all experiments, we use ICNNs with hidden layer sizes [$128$, $128$, $64$, $64$].

\section{Additional Experiments}
\label{sec:additional_experiments}

\begin{figure}[h]
         \centering
         \label{fig:reference-measure}
         \includegraphics[width=1\columnwidth]{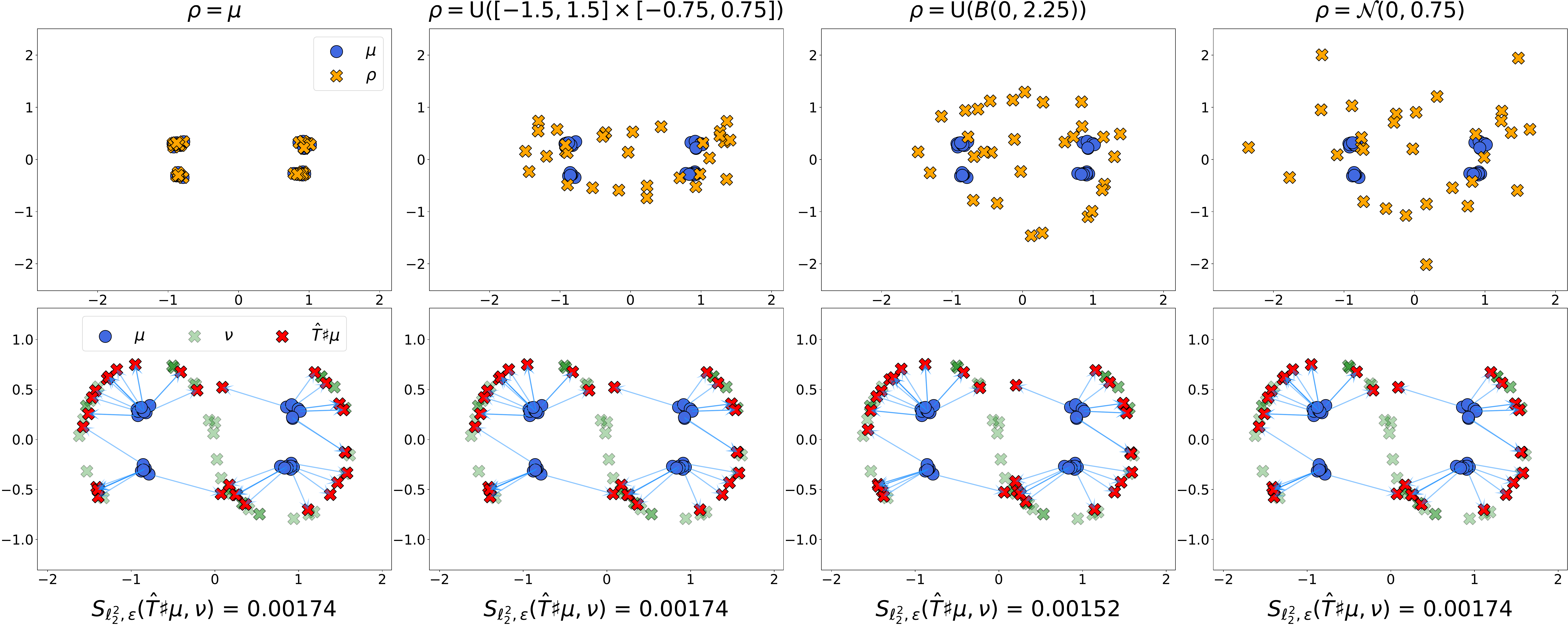}
         \label{fig:reference-measure}
         \caption{Influence of the reference measure $\rho$ when fitting an optimal map between two synthetic measures $\mu, \nu$ for the cost $c(\*x, \*y) = \||\*x - \*y\|_2$ using the induced Monge gap $\mathcal{M}_\mu^1$. The simplest choice is $\rho=\mu$ but by virtue of Proposition \ref{prop:more-do-less}, we can choose any measure $\rho$ such that $\supp(\mu) \subset \supp(\rho)$. Therefore, we estimate a map $\hat{T}$ for 4 reference measures $\rho$ verifying this hypothesis and compare the results: (i) $\rho = \mu$, (ii) $\rho = \mathrm{U}([-1.5, 1.5] \times [-0.75, 0.75])$, (iii) $\rho = \mathrm{U}(B(0, 2.25))$ and (iv) $\rho = \mathcal{N}(0, 0.75)$. For each fitting, we use $W_{\ell_2^2, \varepsilon}$ as the fitting loss and $\lambda_\mathrm{MG} = 1$, and we plot the Sinkhorn divergence $S_{\ell_2^2, \varepsilon}(\hat{T}\sharp\mu, \nu)$ computed on $8,192$ samples from the source and the target measure. We then observe almost identical performances for each $\rho$. In this case, the Monge gap $\mathcal{M}_\rho^1$ seems robust to the choice of the reference measure. We can nevertheless note that the (slightly) best performances are obtained for $\rho = \mathrm{U}(B(0, 2.25))$. The choice of $\rho = \mu$ may not be the best choice, exploiting this track would be an interesting direction for future works.}
\end{figure}

\begin{figure}
         \centering
         \label{fig:fitted-map-receptor}
         \includegraphics[width=.9\columnwidth]{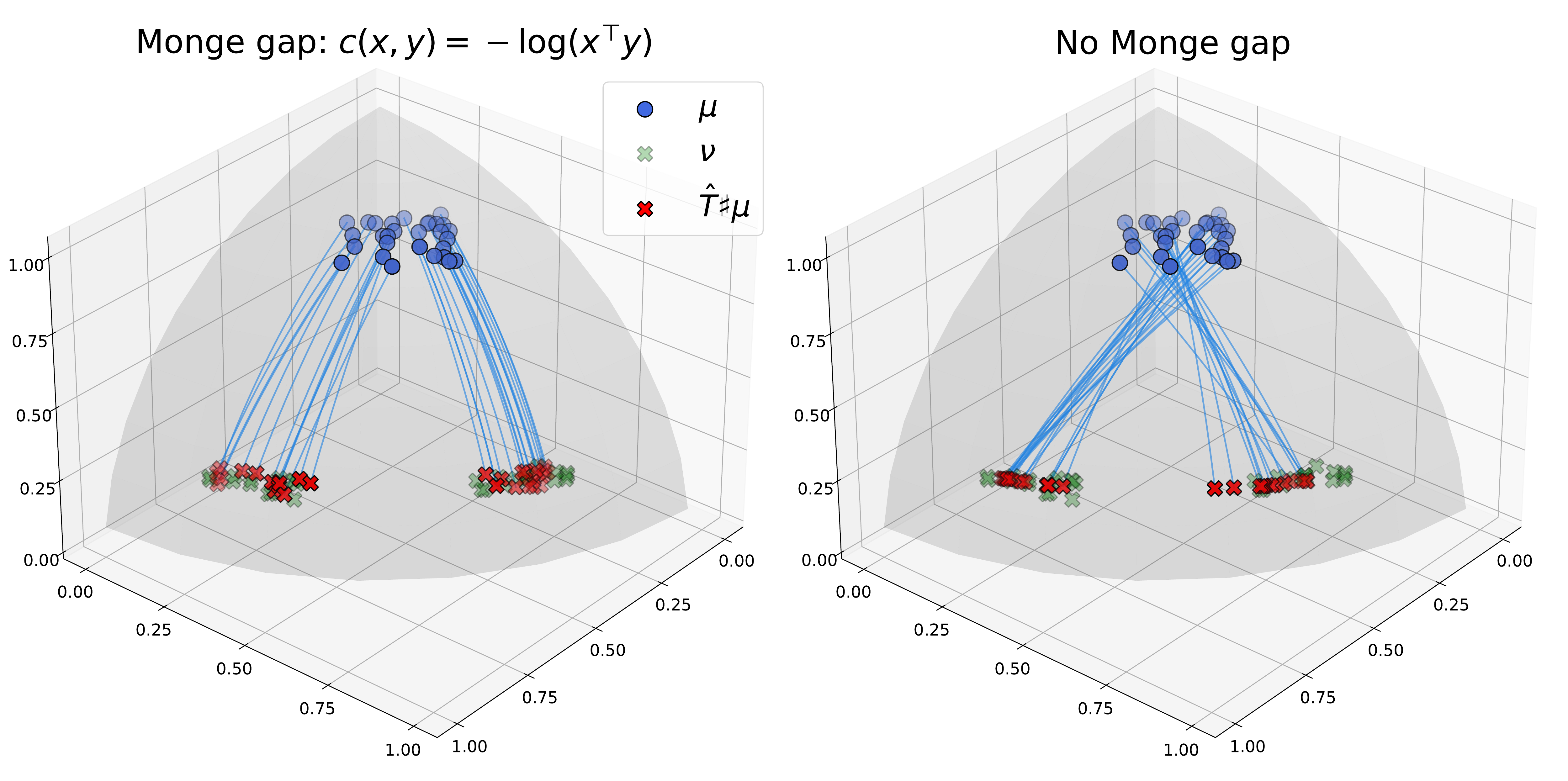}
         \label{fig:fitted-map-receptor}
         \caption{Fitting of 2 transport map between synthetic measures supported on the $2$-sphere. For the left figure, we use the Monge gap instantiated cost $c(\*x, \*y) = - \log(\*x^\top\*y)$ along with $\lambda_\mathrm{MG} = 1$ while we do not use regularizer for the left figure. In both cases, we parameterize the map as $T_\theta = F_\theta / \|F_\theta\|_2$ where $F_\theta$ is an MLP and use $W_{\ell_2^2, \varepsilon}$ as fitting loss.}
\end{figure}

\label{sec:gradient-flow}
\begin{figure}
         \centering
         \label{fig:gradient-flow}
         \includegraphics[width=1\columnwidth]{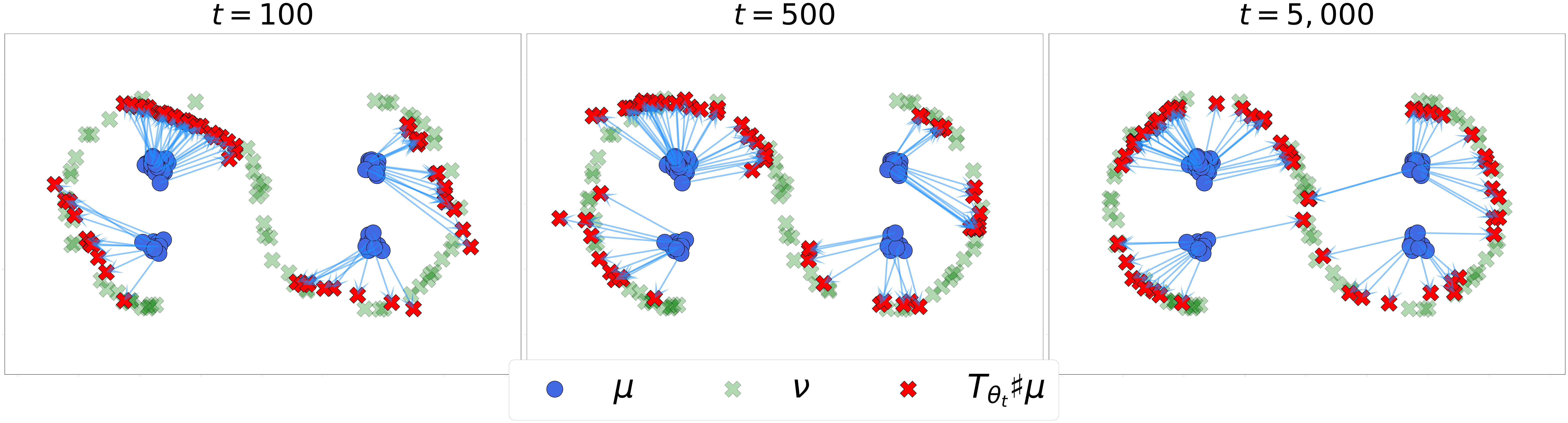}
         \caption{Transport map $(T_{\theta_t})_{t\geq0}$ along the gradient flow of the loss $\mathcal{L}(\theta) = W_{\ell_2^2, \varepsilon}(T_\theta \sharp \mu, \nu) + \lambda_{\mathrm{MG}} \mathcal{M}^1_\mu(T_\theta)$ to fit an optimal map for cost $c(\*x, \*y) = \|\*x - \*y \|_2$ between two synthetic measures $\mu$ and $\nu$. We use $\lambda_{\mathrm{MG}} = 1$. $T_\theta$ is directly paramtrized as a MLP and randomly initialized. We report 3 timestamps of the optimization, at iterations $100$, $500$ and $10,000$. We observe the effect of the Monge gap on the fitting: as the optimization proceeds, the assignment induced by $T_\theta$ tends to respect the "nocrossing" property, as $c$ is a distance.}
\end{figure}

\end{document}